\pdfoutput=1

\documentclass[11pt]{article}

\usepackage{EACL2023}

\usepackage{times}
\usepackage{latexsym}

\usepackage[T1]{fontenc}

\usepackage[utf8]{inputenc}

\usepackage{microtype}

\usepackage{booktabs}
\usepackage{threeparttable}
\usepackage{multirow}
	
\usepackage{color, colortbl}
\definecolor{Gray}{gray}{0.9}

\usepackage{xcolor}

\usepackage{graphicx}
\usepackage{caption}
\usepackage{subcaption}

\usepackage{xr}

\usepackage{soul}

\usepackage{float}

\title{LoRaLay: A Multilingual and Multimodal Dataset for\\ \emph{Lo}ng \emph{Ra}nge and \emph{Lay}out-Aware Summarization}

\author{Laura Nguyen$^{1,3}$ \quad Thomas Scialom$^{2*}$ \quad Benjamin Piwowarski$^3$ \quad Jacopo Staiano$^{4*}$  \\
$^1$reciTAL, Paris, France \quad $^2$Meta AI, Paris, France \\ $^3$Sorbonne Universit\'e, CNRS, ISIR, F-75005 Paris, France  \\ $^4$University of Trento, Italy \\
\\
  \texttt{laura@recital.ai} \quad \texttt{tscialom@fb.com} \\
  \texttt{benjamin.piwowarski@cnrs.fr} \quad \texttt{jacopo.staiano@unitn.it}}

\begin{document}
\maketitle

\def\thefootnote{*}\footnotetext{Work partially done while at reciTAL.}\def\thefootnote{\arabic{footnote}}

\begin{abstract}
Text Summarization is a popular task and an active area of research for the Natural Language Processing community. It requires accounting for long input texts, a characteristic which poses computational challenges for neural models. 
Moreover, real-world documents come in a variety of complex, visually-rich, layouts. This information is of great relevance, whether to highlight salient content or to encode long-range interactions between textual passages. Yet, all publicly available summarization datasets only provide plain text content.
To facilitate research on how to exploit visual/layout information to better capture long-range dependencies in summarization models, we present \textit{LoRaLay}, a collection of datasets for long-range summarization with accompanying visual/layout information. We extend existing and popular English datasets (arXiv and PubMed) with visual/layout information and propose four novel datasets -- consistently built from scholar resources -- covering French, Spanish, Portuguese, and Korean languages.
Further, we propose new baselines merging layout-aware and long-range models -- two orthogonal approaches -- and obtain state-of-the-art results, showing the importance of combining both lines of research.

\end{abstract}

\section{Introduction}
Deep learning techniques have enabled remarkable progress in Natural Language Processing (NLP) in recent years \citep{devlin2018bert, raffel2019exploring, brown2020language}. However, the majority of models, benchmarks, and tasks have been designed for unimodal approaches, i.e. focusing exclusively on a single source of information, namely plain text. While it can be argued that for specific NLP tasks, such as textual entailment or machine translation, plain text is all that is needed, there exist several tasks for which disregarding the visual appearance of text is clearly sub-optimal: in a real-world context (business documentation, scientific articles, etc.), text does not naturally come as a sequence of characters, but is rather displayed in a bi-dimensional space containing rich visual information. 
The layout of e.g. this very paper provides valuable semantics to the reader: in which section are we right now? At the blink of an eye, this information is readily accessible via the salient section title (formatted differently and placed to highlight its role) preceding these words. Just to emphasize this point, \emph{imagine having to scroll this content in plain text to access such information}.

In the last couple of years, the research community has shown a growing interest in addressing these limitations. Several approaches have been proposed to deal with visually-rich documents and integrate layout information into language models, with direct applications to Document Understanding tasks. 
Joint multi-modal pretraining \citep{xu-etal-2021-layoutlmv2, powalski2021going, appalaraju2021docformer} has been key to reach state-of-the-art performance on several benchmarks \citep{jaume2019funsd, gralinski2020kleister, mathew2021docvqa}. Nonetheless, a remaining limitation is that these  (transformer-based) approaches are not suitable for processing long documents, the quadratic complexity of self-attention constraining their use to short sequences. Such models are hence unable to encode global context (e.g. long-range dependencies among text blocks).

Focusing on compressing the most relevant information from long texts to short summaries, the Text Summarization task naturally lends itself to benefit from such global context. Notice that, in practice, the limitations linked to sequence length are also amplified by the lack of visual/layout information in the existing datasets.
Therefore, in this work, we aim at spurring further research on how to incorporate multimodal information to better capture long-range dependencies. 

Our contributions can be summarized as follows:

\begin{itemize}
    \item We extend two popular datasets for long-range summarization, arXiv and PubMed \citep{cohan2018discourse}, by including visual and layout information -- thus allowing direct comparison with previous works;
    \item We release 4 additional layout-aware summarization datasets (128K documents), covering French, Spanish, Portuguese, and Korean languages;
    \item We provide baselines including adapted architectures for multi-modal long-range summarization, and report results showing that (1) performance is far from being optimal; and (2) layout provides valuable information.
\end{itemize}

All the datasets are available on HuggingFace.\footnote{{\tiny\url{https://hf.co/datasets/nglaura/arxivlay-summarization},  \url{https://hf.co/datasets/nglaura/pubmedlay-summarization}, \url{https://hf.co/datasets/nglaura/hal-summarization}, \url{https://hf.co/datasets/nglaura/scielo-summarization}, \url{https://hf.co/datasets/nglaura/koreascience-summarization}}}

\section{Related Work}

\subsection{Layout/Visually-rich Datasets}

Document Understanding covers problems that involve reading and interpreting visually-rich documents (in contrast to plain texts), requiring comprehending the conveyed multimodal information. Hence, several tasks with a central layout aspect have been proposed by the document understanding community. 
\emph{Key Information Extraction} tasks consist in extracting the values of a given set of keys, e.g., the \textit{total amount} in a receipt or the \textit{date} in a form. In such tasks, documents have a layout structure that is crucial for their interpretation. Notable datasets include FUNSD \citep{jaume2019funsd} for form understanding in scanned documents, and SROIE \citep{huang2019icdar2019}, as well as CORD \citep{park2019cord}, for information extraction from receipts.
\citet{gralinski2020kleister} elicit progress on deeper and more complex Key Information Extraction by introducing the Kleister datasets, a collection of business documents with varying lengths, released as PDF files. However, the documents in Kleister often contain single-column layouts, which are simpler than the various multi-column layouts considered in LoRaLay.
\emph{Document VQA} is another popular document understanding task that requires processing multimodal information (e.g., text, layout, font style, images) conveyed by a document to be able to answer questions about a visually rich document (e.g., \textit{What is the date given at the top left of the form?}, \textit{Whose picture is given in this figure?}). The DocVQA dataset \citep{mathew2021docvqa} and InfographicsVQA \citep{mathew2022infographicvqa} are commonly-used VQA datasets that respectively provide industry documents and infographic images, encouraging research on understanding documents with complex interplay of text, layout and graphical elements. 
Finally, to foster research on visually-rich document understanding, \citet{borchmann2021due} introduce the Document Understanding Evaluation (DUE) benchmark, a unified benchmark for end-to-end document understanding, created by combining several datasets. DUE includes several available and transformed datasets for VQA, Key Information Extraction and Machine Reading Comprehension tasks.

\subsection{Existing Summarization Datasets} 

Several large-scale summarization datasets have been proposed to boost research on text summarization systems. \citet{hermann2015teaching} proposed the CNN/DailyMail dataset, a collection of English articles extracted from the CNN and The Daily Mail portals. Each news article is associated with multi-sentence highlights which serve as reference summaries. \citet{scialom2020mlsum} bridge the gap between English and non-English resources for text summarization by introducing MLSum, a large-scale multilingual summarization corpus providing news articles written in French, German, Spanish, Turkish and Russian. 
Going toward more challenging scenarios involving significantly longer documents, the arXiv and PubMed datasets \citep{cohan2018discourse} consist of scientific articles collected from academic repositories, wherein the paper abstracts are used as summaries. To encourage a shift towards building more abstractive summarization models with global content understanding, \citet{sharma2019bigpatent} introduce BIGPATENT, a large-scale dataset made of U.S. patent filings. Here, invention descriptions serve as reference summaries. 

The vast majority of summarization datasets only deal with plain text documents. As opposed to other Document Understanding tasks (e.g., form understanding, visual QA) in which the placement of text on the page and/or visual components are the main source of information needed to find the desired data \citep{borchmann2021due}, text plays a predominant role in document summarization. However, guidelines for summarizing texts -- especially long ones -- often recommend roughly previewing them to break them down into their major sections \citep{toprak2009three, luo2019reading}. 
This suggests that NLP systems might leverage multimodal information in documents.
\citet{miculicich2022document} propose a two-stage method which detects text segments and incorporates this information in an extractive summarization model. \citet{cao2022hibrids} collect a new dataset for long and structure-aware document summarization, consisting of 21k documents written in English and extracted from WikiProject Biography. 

Although not all documents are explicitly organized into clearly defined sections, the great majority contains layout and visual clues (e.g., a physical organization into paragraphs, bigger headings/subheadings) which help structure their textual contents and facilitate reading. Thus, we argue that layout is crucial to summarize long documents. We propose a corpus of more than 345K long documents with layout information. Furthermore, to address the need for multilingual training data \citep{chi2020cross}, we include not only English documents, but also French, Spanish, Portuguese and Korean ones.

\section{Datasets Construction}

Inspired by the way the arXiv and PubMed datasets were built \citep{cohan2018discourse}, we construct our corpus from research papers, with abstracts as ground-truth summaries. As the PDF format allows simultaneous access to textual, visual and layout information, we collect PDF files to construct our datasets, and provide their URLs.\footnote{\scriptsize We make the corpus-construction code publicly available at \url{https://github.com/recitalAI/loralay-datasets}.}

For each language, we select a repository that contains a high number of academic articles (in the order of hundreds of thousands) and provides easy access to abstracts. 
More precisely, we chose the following repositories:
\begin{itemize}
    \item Archives Ouverte HAL (French),\footnote{\scriptsize \url{https://hal.archives-ouvertes.fr/}} an open archive of scholarly documents from all academic fields. As HAL is primarily directed towards French academics, a great proportion of articles are written in French;
    \item SciELO (Spanish and Portuguese),\footnote{\scriptsize \url{https://www.scielo.org/}} an open access database of academic articles published in journal collections from Latin America, Iberian Peninsula and South Africa, and covering a broad range of topics (e.g. agricultural sciences, engineering, health sciences, letters and arts). Languages include English, Spanish, and Portuguese.
    \item KoreaScience (Korean),\footnote{\scriptsize \url{http://www.koreascience.or.kr}} an open archive of Korean scholarly publications in the fields of natural sciences, life sciences, engineering, and humanities and social sciences. Articles are written in English or Korean.
\end{itemize}

Further, we provide enhanced versions of the arXiv and PubMed datasets, respectively denoted as arXiv-Lay and PubMed-Lay, for which layout information is provided.

\subsection{Collecting the Data}

\paragraph{Extended Datasets}

The arXiv and PubMed datasets \cite{cohan2018discourse} contain long scientific research papers extracted from the arXiv and PubMed repositories. We augment them by providing their PDFs, allowing access to layout and visual information. As the abstracts contained in the original datasets are all lowercased, we do not reuse them, but rather extract the raw abstracts using the corresponding APIs.

Note that we were unable to retrieve all the original documents. For the most part, we failed to retrieve the corresponding abstracts, as they did not necessarily match the ones contained in the PDF files (due to \textit{e.g.} PDF-parsing errors). We also found that some PDF files were unavailable, while others were corrupted or scanned documents.\footnote{\scriptsize For more details on this, see Section~\ref{supp-subsec:lost-docs} in the Appendix.} In total, about 39\% (35\%) of the original documents in arXiv (PubMed) were lost.

\subparagraph{arXiv-Lay}

The original arXiv dataset \cite{cohan2018discourse} was constructed by converting the \LaTeX~ files to plain text. To be consistent with the other datasets -- for which \LaTeX~files are not available -- we instead use the PDF files to extract both text and layout elements. For each document contained in the original dataset, we fetch (when possible) the corresponding PDF file using Google Cloud Storage buckets. As opposed to the original procedure, we do not remove tables nor discard sections that follow the conclusion. We retrieve the corresponding abstracts from a metadata file provided by Kaggle.\footnote{\scriptsize \url{https://www.kaggle.com/Cornell-University/arxiv}}

\subparagraph{PubMed-Lay} 

For PubMed, we use the PMC OAI Service\footnote{\scriptsize\url{https://www.ncbi.nlm.nih.gov/pmc/tools/oai/}} to retrieve abstracts and PDF files. 

\paragraph{HAL} 

We use the HAL API\footnote{\scriptsize \url{https://api.archives-ouvertes.fr/docs/search}} to download research papers written in French. To avoid excessively long (e.g. theses) or short (e.g. posters) documents, extraction is restricted to journal and conference papers. 

\paragraph{SciELO}

Using Scrapy,\footnote{\scriptsize \url{https://scrapy.org/}} we crawl the following SciELO collections: Ecuador, Colombia, Paraguay, Uruguay, Bolivia, Peru, Portugal, Spain and Brazil. We download documents written either in Spanish or Portuguese, according to the metadata, obtaining two distinct datasets: SciELO-ES (Spanish) and SciELO-PT (Portuguese).

\paragraph{KoreaScience}

Similarly, we scrape the KoreaScience website to extract research papers. We limit search results to documents whose publishers’ names contain the word \emph{Korean}. This rule was designed after sampling documents in the repository, and is the simplest way to get a good proportion of papers written in Korean.\footnote{\scriptsize For further details, see Section~\ref{supp-sec:koreascience-extraction} in the Appendix.} Further, search is restricted to papers published between 2012 and 2021, as recent publications are more likely to have digital-born, searchable PDFs. Finally, we download the PDF files of documents that contain an abstract written in Korean. 

\subsection{Data Pre-processing}

For each corpus, we use the 95th percentile of the page distribution as an upper bound to filter out documents with too many pages, while the 5th (1st for HAL and SciELO) percentile of the summary length distribution is used as a minimum threshold to remove documents whose abstracts are too short. As our baselines do not consider visual information, we only extract text and layout from the PDF files. Layout is incorporated by providing the spatial position of each word in a document page image, represented by its bounding box $(x_0, y_0, x_1, y_1)$, where $(x_0, y_0)$ and $(x_1, y_1)$ respectively denote the coordinates of the top-left and bottom-right corners. Using the PDF rendering library Poppler\footnote{\scriptsize \url{https://poppler.freedesktop.org/}}, text and word bounding boxes are extracted from each PDF, and the sequence order is recovered based on heuristics around the document layout (e.g., tables, columns). Abstracts are then removed by searching for exact matches; when no exact match is found, we use \texttt{fuzzysearch}\footnote{\scriptsize \url{https://pypi.org/project/fuzzysearch/}} and \texttt{regex}\footnote{\scriptsize \url{https://pypi.org/project/regex/}} to find near matches.\footnote{\scriptsize We use a maximum Levenshtein distance of 20 with fuzzysearch, and a maximum number of errors of 3 with regex.} For the non-English datasets, documents might contain several abstracts, written in different languages. To avoid information leakage, we retrieve the abstract of each document in every language available -- according to the API for HAL or the websites for SciELO and KoreaScience -- and remove them using the same strategy as for the main language. In the case an abstract cannot be found, we discard the document to prevent any unforeseen leakage. The dataset construction process is illustrated in Section~\ref{supp-sec:datasets-construction} in the Appendix. 

\subsection{Datasets Statistics}

The statistics of our proposed datasets, along with those computed on existing summarization datasets of long documents \citep{cohan2018discourse, sharma2019bigpatent} are reported in Table~\ref{tab:datasets-stats}. We see that document lengths are comparable or greater than for the arXiv, PubMed and BigPatent datasets.  

For arXiv-Lay and PubMed-Lay, we retain the original train/validation/splits and try to reconstruct them as faithfully to the originals as possible. For the new datasets, we order documents based on their publication dates and provide splits following a chronological ordering. For HAL and KoreaScience, we retain 3\% of the articles as validation data, 3\% as test, and the remaining as training data. To match the number of validation/test documents in HAL and KoreaScience, we split the data into 90\% for training, 5\% for validation and 5\% for test, for both SciELO datasets.

\begin{table}
\centering
\small
\resizebox{\linewidth}{!}{%
\begin{tabular}{ccccc}
\hline
             \multirow{3}{*}{\textbf{Dataset}} & \textbf{\# Docs} & \textbf{Mean}         & \textbf{Mean}  \\
             &                  & \textbf{Article}        & \textbf{Summary} \\
             &                  & \textbf{Length}        & \textbf{Length} \\
             \hline
arXiv \citep{cohan2018discourse}      & 215,913   & 3,016 & 203  \\
PubMed \citep{cohan2018discourse}     & 133,215   & 4,938 & 220   \\
BigPatent \citep{sharma2019bigpatent} & 1,341,362 & 3,572 & 117 \\
\midrule
arXiv-Lay       & 130,919 & 7,084 & 125 \\
PubMed-Lay      & 86,668  & 4,038 & 144 \\
HAL          & 46,148  & 4,543 & 134 \\
SciELO-ES    & 23,170  & 4,977 & 172 \\
SciELO-PT    & 21,563  & 6,853 & 162 \\
KoreaScience & 37,498  & 3,192 &  95 \\
\hline
\end{tabular}
}
\caption{Datasets statistics. Article and summary lengths are computed in words. For KoreaScience, words are obtained via white-space tokenization. Difference between arXiv and arXiv-Lay is due to the fact that we retain the whole document, while \citet{cohan2018discourse} truncate it after the conclusion.}
\label{tab:datasets-stats}
\end{table}

\begin{table*}[]
\centering
\small 
\begin{threeparttable}
\begin{tabular}{crrrrrrrrrr}
\hline
\multicolumn{1}{c}{\multirow{2}{*}{\textbf{Dataset}}} & & \multicolumn{3}{c}{\textbf{Instances}} &                                                  & \multicolumn{2}{c}{\textbf{Input Length}} & & \multicolumn{2}{c}{\textbf{Output Length}} \\
\multicolumn{1}{l}{} & & \multicolumn{1}{c}{\textbf{Train}} & \multicolumn{1}{c}{\textbf{Dev}} & \multicolumn{1}{c}{\textbf{Test}} & & \textbf{Median}         & \textbf{90\%-ile}        & & \textbf{Median}         & \textbf{90\%-ile}         \\
\hline
arXiv \citep{cohan2018discourse}  & & 203,037 & 6,436 & 6,440 & & 6,151 & 14,405 & & 171 & 352 \\
PubMed \citep{cohan2018discourse} & & 119,924 & 6,633 & 6,658 & & 2,715 &  6,101 & & 212 & 318 \\
\midrule
arXiv-Lay               & & 122,189                   & 4,374                   & 4,356      &              &  6,225         & 12,541           &   & 150          & 249     \\
PubMed-Lay              & & 78,234                    & 4,084                   & 4,350      &              & 
3,761        & 7,109           &  & 182           & 296              \\
HAL                  & & 43,379                    & 1,384                   & 1,385      &              & 4,074          & 8,761           & & 179            & 351              \\
SciELO-ES            & & 20,853                    & 1,158                   & 1,159      &               & 4,859          & 8,519           & & 226            & 382              \\
SciELO-PT            & & 19,407                    & 1,078                   & 1,078      &               &  6,090          & 9,655           &  & 239                &  374                \\
KoreaScience         & & 35,248                    & 1,125                   & 1,125      &              &  2,916          &  5,094           &  & 219           & 340   \\
\hline 
\end{tabular}
\end{threeparttable}
\caption{Datasets splits and statistics. Input and output lengths are computed in tokens, obtained using Pegasus and MBART-50's tokenizers for the English and non-English datasets, respectively.}
\end{table*}

\section{Experiments}

\subsection{Models}

For reproducibility purposes, we make the models implementation, along with the fine-tuning and evaluation scripts, publicly available.\footnote{\scriptsize \url{https://github.com/recitalAI/loralay-modeling}}

We do not explore the use of visual information in long document summarization, as the focus is on evaluating baseline performance using state-of-the-art summarization models augmented with layout information. While visual features might provide a better understanding of structures such as tables and figures, we do not expect substantial gains with respect to layout-aware models. Indeed, the information provided in figures (i.e., information that cannot be captured by layout or text) are commonly described in the caption or related paragraphs. 

\paragraph{Text-only models with standard input size}
    
We use Pegasus \citep{zhang2020pegasus} as a text-only baseline for arXiv-Lay and PubMed-Lay. Pegasus is an encoder-decoder model pre-trained using gap-sentences generation, making it a state-of-the-art model for abstractive summarization.
For the non-English datasets, we rely on a finetuned MBART as our baseline. MBART \citep{liu-etal-2020-multilingual-denoising} is a multilingual sequence-to-sequence model pretrained on large-scale monolingual corpora in many languages using the BART objective \citep{lewis2019bart}. We use its extension, MBART-50 \citep{tang2020multilingual},\footnote{\scriptsize For the sake of clarity, we refer to MBART-50 as MBART.} which is created from the original MBART by extending its embeddings layers and pre-training it on a total of 50 languages. 
Both Pegasus and MBART are limited to a maximum sequence length of 1,024 tokens, which is well below the median length of each dataset.
\vspace{-0.1cm}
\paragraph{Layout-aware models with standard input size}

We introduce layout-aware extensions of Pegasus and MBART, respectively denoted as Pegasus+Layout and MBART+Layout. Following LayoutLM \citep{xu2020layoutlm}, which is state-of-the-art on several document understanding tasks \citep{jaume2019funsd, huang2019icdar2019, harley2015evaluation}, each token bounding box coordinates $(x_0, y_0, x_1, y_1)$ is normalized into an integer in the range [0, 1000]. Spatial positions are encoded using four embedding tables, namely two for the coordinate axes ($x$ and $y$), and the other two for the bounding box size (width and height). The layout representation of a token is formed by summing the resulting embedding representations
The final representation of a token is then obtained through point-wise summation of its textual, 1D-positional and layout embeddings.

\paragraph{Long-range, text-only models}

To process longer sequences, we leverage BigBird \citep{zaheer2020big}, a sparse-attention based Transformer which reduces the quadratic dependency to a linear one. For arXiv-Lay and PubMed-Lay, we initialize BigBird from Pegasus \citep{zaheer2020big} and for the non-English datasets, we use the weights of MBART. The resulting models are referred to as BigBird-Pegasus and BigBird-MBART. For both models, BigBird sparse attention is used only in the encoder. Both models can handle up to 4,096 inputs tokens, which is greater than the median length in PubMed-Lay, HAL and KoreaScience. 

\paragraph{Long-range, layout-aware models}

We also include layout information in long-range text-only models. Similarly to layout-aware models with standard input size, we integrate layout information into our long-range models by encoding each token's spatial position in the page. The resulting models are denoted as BigBird-Pegasus+Layout and BigBird-MBART+Layout.

\paragraph{Additional State-of-the-Art Baselines}

We further consider additional state-of-the-art baselines for summarization: i) the text-only T5 \citep{raffel2019exploring} with standard input size, ii) the long-range Longformer-Encoder-Decoder (LED) \citep{beltagy2020longformer}, and iii) the layout-aware, long-range LED+Layout, which we implement similarly to the previous layout-aware models. \\

\subsection{Implementation Details}

We initialize our Pegasus-based and MBART-based models with, respectively, the google/pegasus-large and facebook/mbart-large-50 checkpoints shared through the Hugging Face Model Hub. As for T5 and LED, we use the weights from t5-base and allenai/led-base-16384, respectively.\footnote{\scriptsize The large versions of T5 and LED did not fit into GPU due to their size.}

Following \citet{zhang2020pegasus} and \citet{zaheer2020big}, we fine-tune our models up to 74k (100k) steps on arXiv-Lay (PubMed-Lay). On HAL, the total number of steps is set to 100k, while it is decreased to 50k for the other non-English datasets.\footnote{\scriptsize We tested different values for the number of steps (10k, 25k, 50k, 100k) and chose the one that gave the best validation scores for MBART.} 
For each model, we select the checkpoint with the best validation loss. For Pegasus and MBART models, inputs are truncated at 1,024 tokens. For BigBird-Pegasus models, we follow \citet{zaheer2020big} and set the maximum input length at 3,072 tokens. As the median input length is much greater in almost every non-English dataset, we increase the maximum input length to 4,096 tokens for BigBird-MBART models. Output length is restricted to 256 tokens for all models, which is enough to fully capture at least 50\% of the summaries in each dataset.

For evaluation, we use beam search and report a single run for each model and dataset. Following \citet{zhang2020pegasus, zaheer2020big}, we set the number of beams to 8 for Pegasus-based models, and 5 for BigBird-Pegasus-based models. For the non-English datasets, we set it to 5 for all models, for fair comparison. For all experiments, we use a length penalty of 0.8. For more implementation details, see Section~\ref{supp-sec:implementation-details} in the Appendix.

\section{Results and Discussion}

\subsection{General Results}

\begin{table*}[ht]
\centering
\small
\begin{tabular}{lccc}
\toprule
\multicolumn{1}{c}{\textbf{Model}} & \textbf{\# Params} & \shortstack{\textbf{arXiv/} \\ \textbf{arXiv-Lay}} & \shortstack{\textbf{PubMed/} \\ \textbf{PubMed-Lay}} \\  
\midrule
 \rowcolor{Gray} Pegasus \citep{zhang2020pegasus} & 568M & 38.83 & 41.34 \\
 \rowcolor{Gray} BigBird-Pegasus \citep{zaheer2020big} & 576M & 41.77 & 42.33\\
\hline
T5 \citep{raffel2019exploring}                     & 223M & 37.90 & 39.23 \\ 
LED \citep{beltagy2020longformer}                   & 161M & 40.74 & 41.54 \\ 
LED+Layout             & 165M & 40.96 & 41.83 \\[0.7mm]
Pegasus                & 568M & 39.07 & 39.75 \\ 
Pegasus+Layout         & 572M & 39.25 & 39.85 \\
BigBird-Pegasus        & 576M & 39.59 & 41.09 \\
BigBird-Pegasus+Layout & 581M & \textbf{41.15} & \textbf{42.05} \\
\bottomrule
                            
\end{tabular}
\caption{ROUGE-L scores on arXiv-Lay and PubMed-Lay. Reported results obtained by Pegasus and BigBird-Pegasus on the original arXiv and PubMed are reported with a gray background. The best results obtained on arXiv-Lay and PubMed-Lay are denoted in bold.}
\label{tab:rl-scores-arxiv-pubmed}
\end{table*}

\begin{table*}[ht]
\centering
\small
\begin{tabular}{lccccc}
\toprule
\multicolumn{1}{c}{\textbf{Model}} & \textbf{\# Params} & \shortstack{\textbf{HAL} \\ (fr)} & \shortstack{\textbf{SciELO-ES} \\ (es)} & \shortstack{\textbf{SciELO-PT} \\ (pt)} & \shortstack{\textbf{KoreaScience} \\ (ko)} \\  
\midrule
MBART                  & 610M & 42.00 & 36.55 & 36.42 & 16.94 \\
MBART+Layout           & 615M & 41.67 & 37.47 & 34.37 & 14.98 \\
BigBird-MBART          & 617M & 45.04 & 37.76 & 39.63 & 18.55 \\
BigBird-MBART+Layout   & 621M & \textbf{45.20} & \textbf{40.71} & \textbf{40.51} & \textbf{19.95} \\
\bottomrule
                            
\end{tabular}
\caption{ROUGE-L scores on the non-English datasets. The best results for each dataset are reported in bold. }
\label{tab:rl-scores-multilingual}
\end{table*}

\begin{table}[ht]
\centering
\small
\begin{tabular}{crrr}
\toprule
\textbf{Dataset} & \textbf{Train} & \textbf{Validation} & \textbf{Test} \\
\midrule
HAL (fr)                                 & 90.72                     & 90.54                          & 85.84                    \\
SciELO-ES (es)                            & 84.86                     & 84.28                          & 84.90                    \\
SciELO-PT (pt)                           & 90.95                     & 90.58                          & 91.96                    \\
KoreaScience (ko) & 73.53                     & 70.26                          & 68.78       \\            
\bottomrule
\end{tabular}
\caption{Percent confidence obtained for the main language, for each dataset split.}
\label{table:percentage-main-lang}
\end{table}

\begin{figure*}[ht]
    \centering
    \begin{subfigure}[b]{0.32\textwidth}
        \includegraphics[width=1\textwidth]{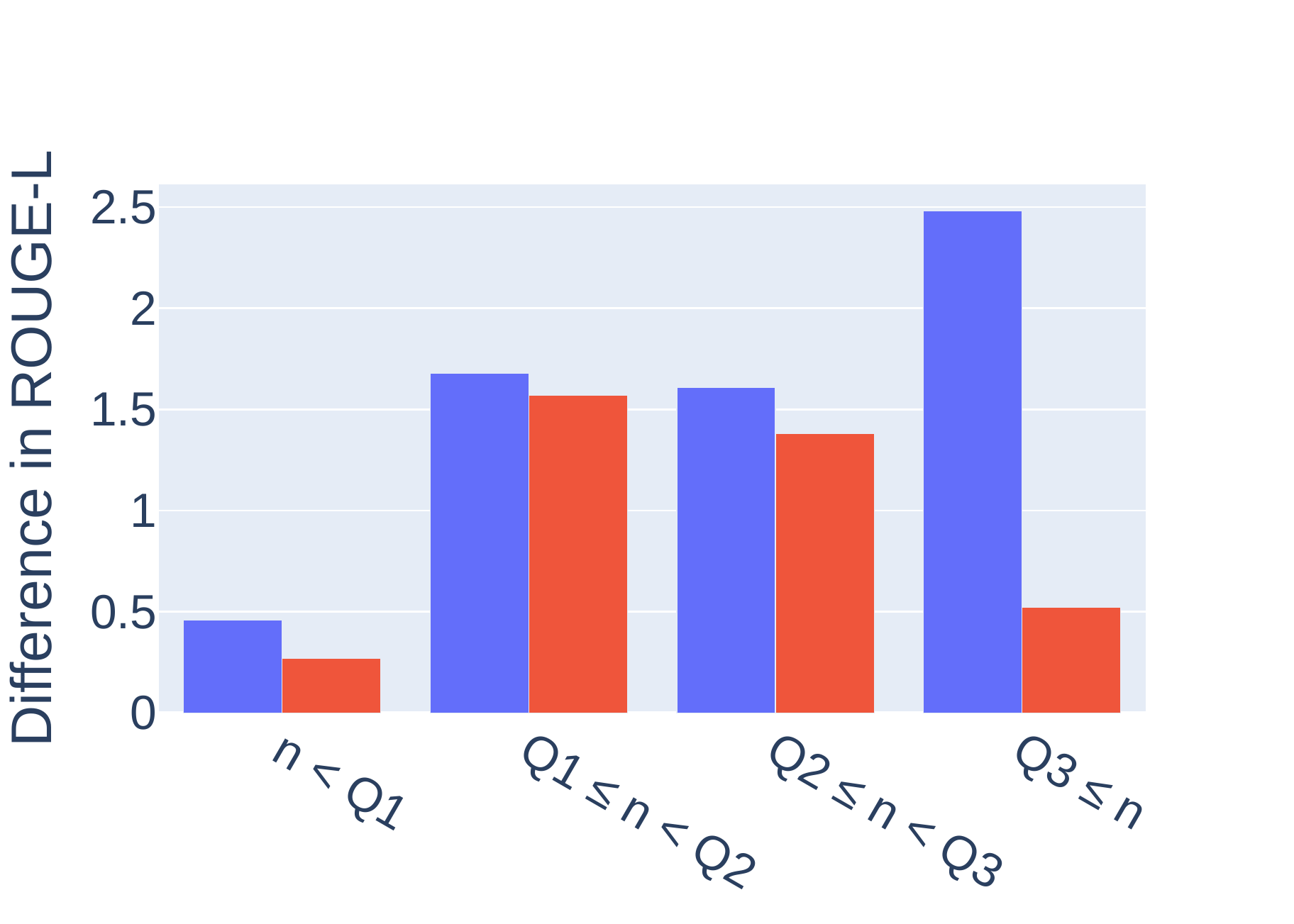}
        \caption{Article length}
    \end{subfigure}
    \begin{subfigure}[b]{0.32\textwidth}
        \includegraphics[width=\textwidth]{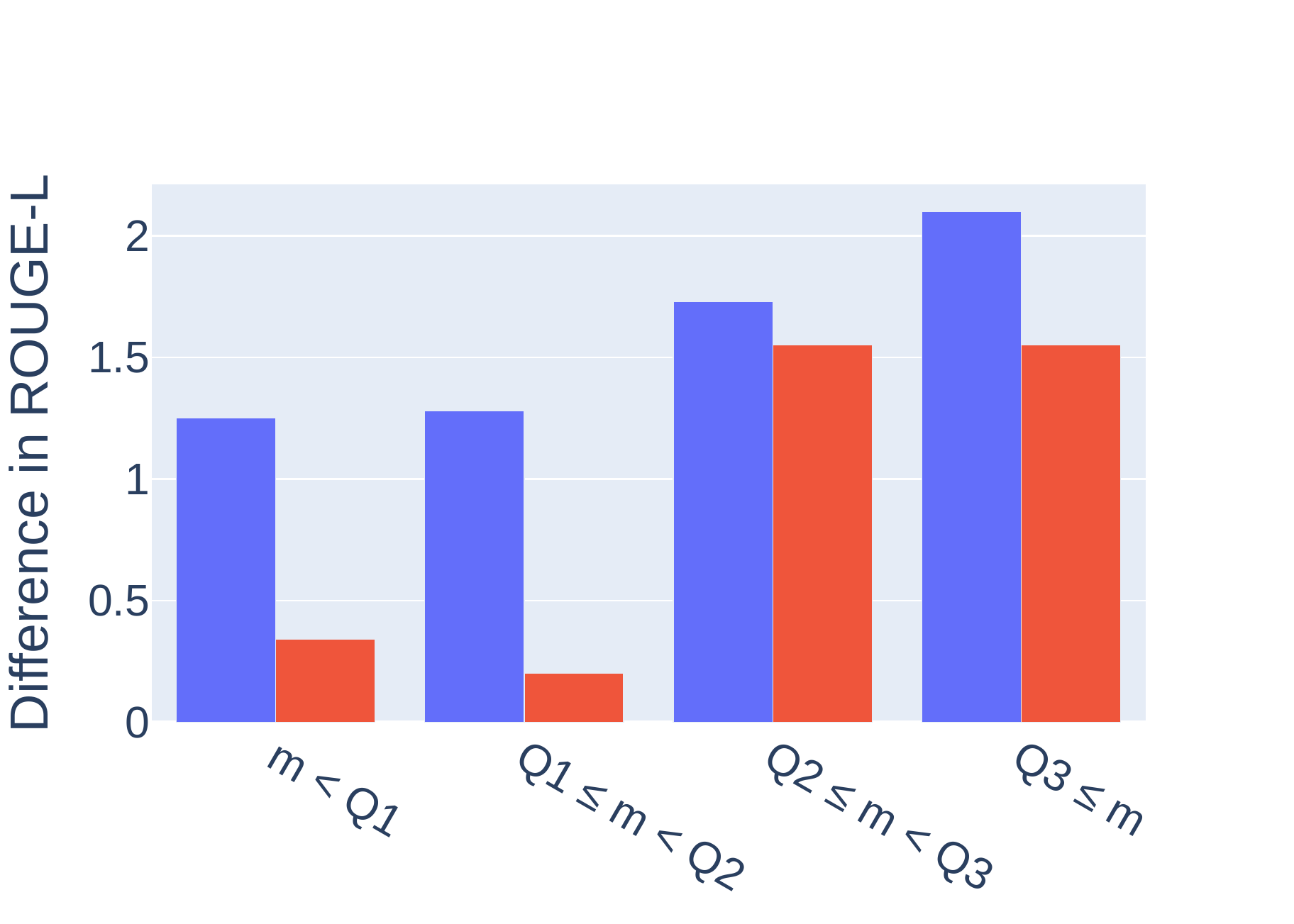}
        \caption{Summary length}
    \end{subfigure}
    \begin{subfigure}[b]{0.32\textwidth}
        \includegraphics[width=\textwidth]{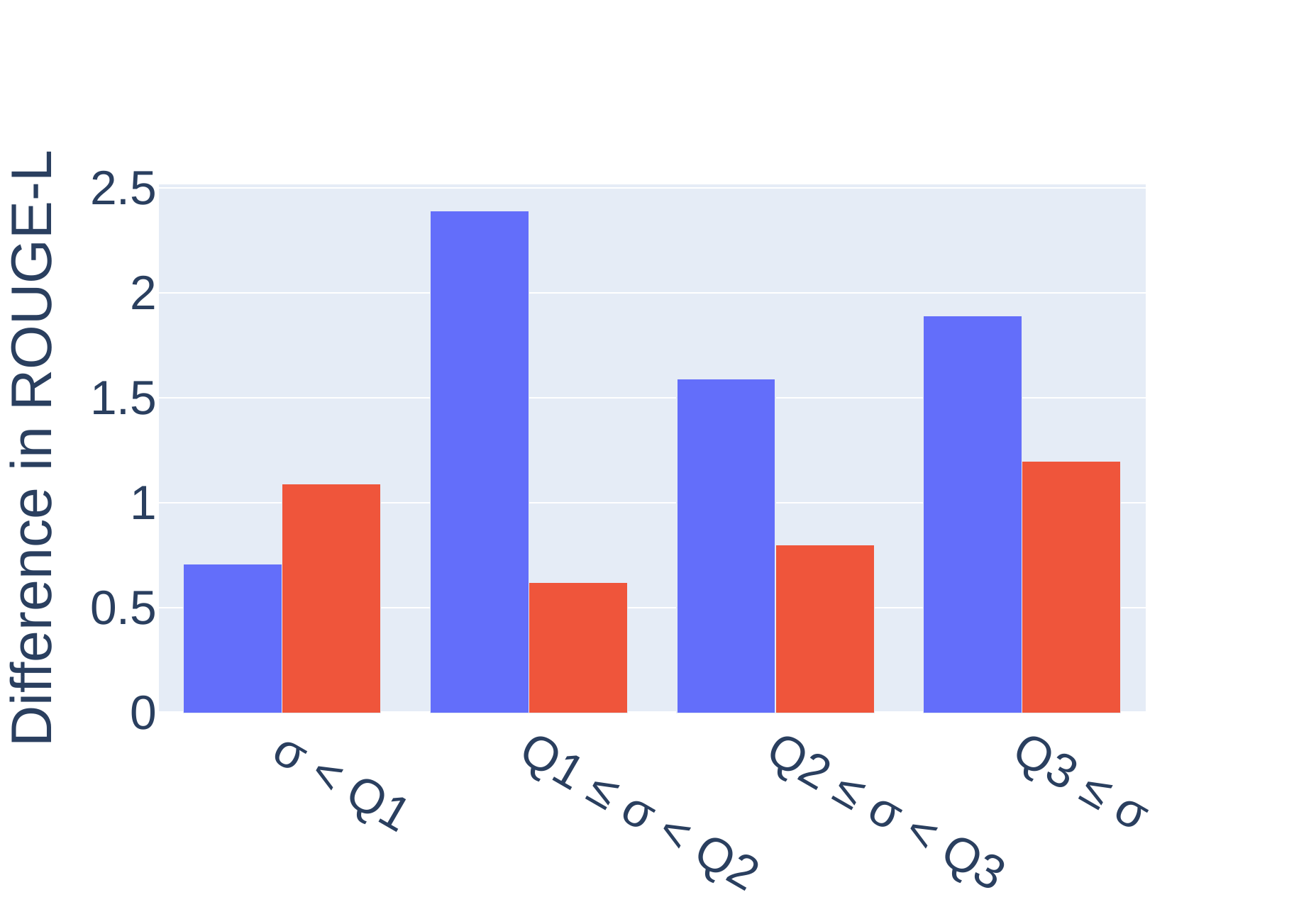}
        \caption{$\sigma$ of bounding box height}
    \end{subfigure}
    \caption{Benefit of using layout on arXiv-Lay (blue) and PubMed-Lay (red), defined as the difference in ROUGE-L scores between BigBird-Pegasus+Layout and BigBird-Pegasus. For each dataset, quartiles are calculated from the distributions of article lengths (a), summary lengths (b) and variance in the height of the bounding boxes (c). ROUGE-L scores are then computed per quartile range, and averaged over each range.}
    \label{fig:analysis-quartiles}
\end{figure*}

In Table~\ref{tab:rl-scores-arxiv-pubmed}, we report the ROUGE-L scores obtained on arXiv and PubMed datasets (reported by \citet{zaheer2020big}), as well as on the corresponding layout-augmented counterparts we release. \footnote{\scriptsize For detailed results, please refer to Section~\ref{supp-sec:detailed-results} in the Appendix.}
On arXiv-Lay and PubMed-Lay, we observe that, while the addition of layout to Pegasus does not improve the ROUGE-L scores, there are gains in integrating layout information into BigBird-Pegasus. To assess whether these gains are significant, we perform significance analysis at the 0.05 level using bootstrap, and estimate a ROUGE-L threshold that predicts when improvements are significant. ROUGE-L improvements between each pair of models are reported in Table~\ref{supp-tab:rouge-l-improvements} in the appendix. On arXiv-Lay, we compute a threshold of 1.48 ROUGE-L, showing that BigBird-Pegasus+Layout significantly outperforms all Pegasus-based models. In particular, we find a 1.56 ROUGE-L improvement between BigBird-Pegasus and its layout-augmented counterpart, demonstrating that the addition of layout to long-range modeling significantly improves summarization. On PubMed-Lay, we compute a threshold of 1.77. Hence, the 0.96 ROUGE-L improvement from BigBird-Pegasus to its layout-augmented counterpart is not significant. However, the variance in font sizes in PubMed-Lay is much smaller compared to arXiv-Lay (see Table~\ref{supp-tab:quartiles} in the appendix), reflecting an overall more simplistic layout. Therefore, we argue that layout integration has a lesser impact in PubMed-Lay, which can explain the non-significance of results. In addition, we find that BigBird-Pegasus significantly outperforms Pegasus and Pegasus+Layout only when augmented with layout, with an improvement of, respectively, 2.3 and 2.2 points. This demonstrates the importance of combining layout and long-range modeling. 

While T5 and LED obtain competitive results, we find that the gain in adding layout to LED is minor. However, the models we consider have all been pre-trained only on plain text. As a result, the layout representations are learnt from scratch during fine-tuning. Similarly to us, \citet{borchmann2021due} show that their layout-augmented T5 does not necessarily improve the scores, and that performance is significantly enhanced only when the model has been pre-trained on layout-rich data.  

Further, we observe, for both Pegasus and BigBird-Pegasus, a drop in performance w.r.t. the scores obtained on the original datasets. This can be explained by two factors. First, our extended datasets contain less training data due to the inability to process all original documents. Secondly, the settings are different: while the original arXiv and PubMed datasets contain clear discourse information (e.g., each section is delimited by markers) obtained from \LaTeX~ files, documents in our extended versions are built by parsing raw PDF files. Therefore, the task is more challenging for text-only baselines, as they have no access to the discourse structure of documents, which further underlines the importance of taking the structural information, brought by visual cues, into account.

Table \ref{tab:rl-scores-multilingual} presents the ROUGE-L scores reported on the non-English datasets. 
On HAL, we note that BigBird-MBART does not benefit from layout. After investigation, we hypothesize that this is due to the larger presence of single-column and simple layouts, which makes layout integration less needed.
On both SciELO datasets, we notice that combining layout with long-range modeling brings substantial improvements over MBART.
Further, we find that the plain-text BigBird models do not improve over the layout-aware Pegasus and MBART on arXiv-Lay and SciELO-ES, demonstrating that simply capturing more context does not always suffice. 
Regarding performance on KoreaScience, we can see a significant drop in performance for every model w.r.t the other non-English datasets. At first glance, we notice a high amount of English segments (e.g., tables, figure captions, scientific concepts) in documents in KoreaScience. To investigate this, we use the cld2 library\footnote{\scriptsize\url{https://github.com/GregBowyer/cld2-cffi}} to detect the language in each non-English document. We consider the percent confidence of the top-1 matching language as an indicator of the presence of the main language (i.e., French, Spanish, Portuguese or Korean) in a document, and average the results to obtain a score for the whole dataset. Table~\ref{table:percentage-main-lang} reports the average percent confidence obtained on each split, for each dataset. We find that the percentage of text written in the main language in KoreaScience (i.e., Korean) is smaller than in other datasets. As the MBART-based models expect only one language in a document (the information is encoded using a special token), we claim the strong presence of non-Korean segments in KoreaScience causes them to suffer from interference problems. Therefore, we highlight that KoreaScience is a more challenging dataset, and we hope our work will boost research on better long-range, multimodal \textit{and} multilingual models.

Overall, results show a clear benefit of integrating layout information for long document summarization. 

\subsection{Human Evaluation}

\begin{table}[ht]
\centering
\small
\begin{tabular}{lccccc}
\toprule
\textbf{Metric}        & \textbf{BigBird} & \textbf{BigBird+Layout} \\ 
\midrule
Precision \%    &    35.15 \scriptsize{(0.81)}            &      37.51 \scriptsize{(0.70)}                   \\
Recall \%       &    28.07 \scriptsize{(0.73)}             &     33.59 \scriptsize{(0.86)}                   \\
Coherence     &     3.80 \scriptsize{(0.38)}             &      3.75 \scriptsize{(0.62)}                   \\ 
Fluency       &     4.48 \scriptsize{(0.03)}             &      4.34 \scriptsize{(0.16)}                   \\
Overlap \%     &    8.77 \scriptsize{(0.24)}             &     7.49 \scriptsize{(0.36)}                    \\ 
Flow \%             &   30.75 \scriptsize{(0.68)}          &    33.02 \scriptsize{(0.71)}                     \\
\bottomrule
\end{tabular}
\caption{Average human judgement scores obtained by comparing gold-truth abstracts and summaries generated by BigBird and BigBird+Layout from 50 documents sampled from arXiv-Lay and HAL. Inter-rater agreement	is computed using Krippendorff's alpha coefficient, and enclosed between parentheses.}
\label{tab:human-eval-scores}
\end{table}

To gain more insight into the effect of document layout for summarizing long textual content, we conduct a human evaluation of summaries generated by BigBird-Pegasus/BigBird-MBART and their layout-aware counterparts. We choose the BigBird-based models over the LED ones, as the gain in augmenting BigBird with layout is much more apparent. We evenly sample 50 documents from arXiv-Lay and HAL test sets, filtering documents by their topics (computer science) to match the judgment capabilities of the three human annotators. We design an evaluation interface (see Section~\ref{supp-sec:human-eval} in the appendix). For each sentence $s_i$ in the generated summary, we ask the annotators to highlight the relevant tokens in $s_i$, along with the equivalent parts in the ground-truth abstract (denoted $h_i$). Further, we ask them to rate the summary in terms of coherence and fluency, on a scale of 0 to 5, following the DUC quality guidelines \citep{dang2005overview}. Finally, annotators are asked to penalize summaries with hallucinated facts. The highlighting process allows us to compute precision and recall as the percentage of highlighted information in the generated summary and the ground-truth abstract, respectively. Moreover, we can compute an overlap ratio as the percentage of highlighted information that appears several times in the generated summary. Lastly, we calculate a flow percentage that evaluates how well the order of the ground-truth information is preserved by computing the percentage of times where the highlighted text $h_i$ in the gold summary for one generated sentence $s_i$ follows the highlighted text $h_{i-1}$ for the previous sentence $s_{i-1}$ (i.e. where any token from $h_i$ occurs after a token in $h_{i-1}$).
Table~\ref{tab:human-eval-scores} reports the scores for each metric and model, averaged over all 50 documents, along with inter-rater agreements, computed using Krippendorff's alpha coefficient. We find that adding layout to the models significantly improves precision and recall, results in less overlap (repetition), and is more in line with the ground truth order. Further, annotators did not encounter any hallucinated fact in the 50 generated summaries. To conclude, reported results show that human annotators strongly agree that adding layout generates better summaries, further validating our claim that layout provides vital information for summarization tasks.

\subsection{Case Studies}

To have a better understanding of the previous results, we focus on uncovering the cases in which layout is most helpful. To this end, we identify features that relate to the necessity of having layout: 1) article length, as longer texts are intuitively easier to understand with layout, 2) summary length, as longer summaries are likely to cover more salient information, and 3) variance in font sizes (using the height of the bounding boxes), and, as such, the complexity of the layout.
The benefit of using layout is measured as the difference in ROUGE-L scores between BigBird-Pegasus+Layout and its purely textual counterpart, on arXiv-Lay and PubMed-Lay. We compute quartiles from the distributions of article lengths, ground-truth summary lengths, and variance in the height of bounding boxes.\footnote{\scriptsize The quartiles are provided in Appendix~\ref{supp-sec:analysis-layout}.} Based on the aforementioned factors, the scores obtained by each model are then grouped by quartile range, and averaged over each range, see Figure~\ref{fig:analysis-quartiles}. On arXiv-Lay, we find that layout brings most improvement when dealing with the 25\% longest documents and summaries, while, for both datasets, layout is least beneficial for the shortest documents and summaries. These results corroborate our claim that layout can bring important information about long-range context. Concerning the third factor, we see, on PubMed-Lay, that layout is most helpful for documents that have the widest ranges of font sizes, showcasing the advantage of using layout to capture salient information. 

\vspace{-0.05cm}
\section{Limitations and Risks}

The proposed corpus is limited to a single domain, that of scientific literature. Such limitation arguably extends also to the layout diversity of documents. In terms of risks, we acknowledge the presence of Personally Identifiable Information such as author names and affiliations; nonetheless, such information is already voluntarily made public by the authors themselves. 
\vspace{-0.05cm}
\section{Conclusion}

We have presented LoRaLay, a set of large-scale datasets for long-range and layout-aware text summarization. LoRaLay provides the research community with 4 novel multimodal corpora covering French, Spanish, Portuguese, and Korean languages, built from scientific articles. Furthermore, it includes additional layout and visual information for existing long-range summarization datasets (arXiv and PubMed). We provide adapted architectures merging layout-aware and long-range models, and show the importance of layout information in capturing long-range dependencies.

\section{Acknowledgements}

We thank the reviewers for their insightful comments. This work is supported by the Association Nationale de la Recherche et de la Technologie (ANRT) under CIFRE grant N2020/0916. It was partially performed using HPC resources from GENCI-IDRIS (Grant 2021-AD011011841).

\bibliographystyle{acl_natbib}
\bibliography{anthology, custom}

\clearpage 
\appendix 

\begin{center}{
\Large
\textbf{LoRaLay: A Multilingual and Multimodal Dataset for \emph{Lo}ng \emph{Ra}nge and \emph{Lay}out-Aware Summarization -- Appendix}}
\end{center}

\section{Datasets Construction}
\label{supp-sec:datasets-construction}

\begin{figure}[ht]
\centering
    \includegraphics[width=0.45\textwidth]{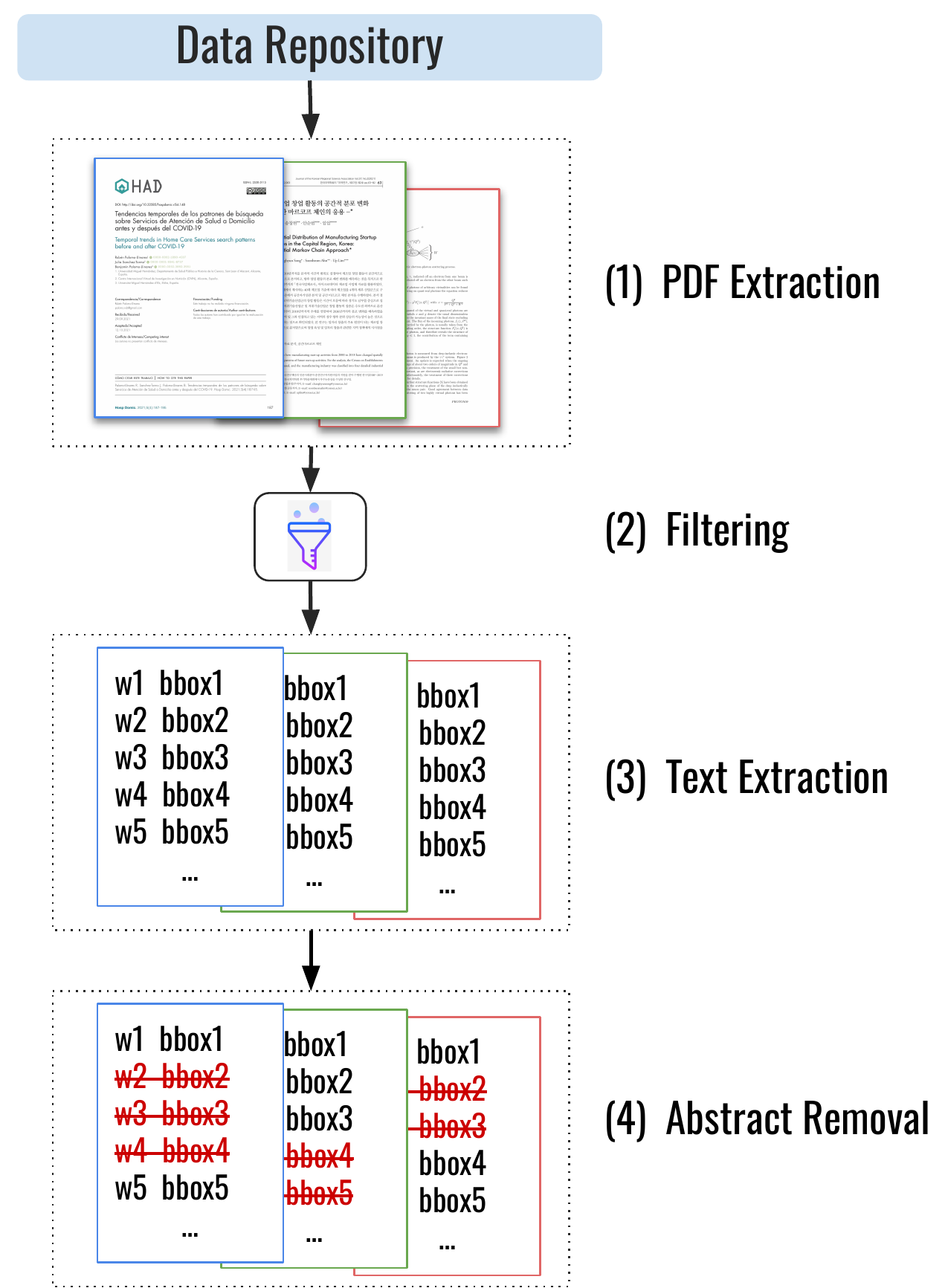}
  \caption{Dataset Construction Process.}
  \label{supp-fig:dataset-construction}
\end{figure}

\subsection{Extended Datasets -- Lost Documents}
\label{supp-subsec:lost-docs}

\begin{figure}[H]
  \begin{subfigure}[b]{0.4\textwidth}
    \includegraphics[width=\textwidth]{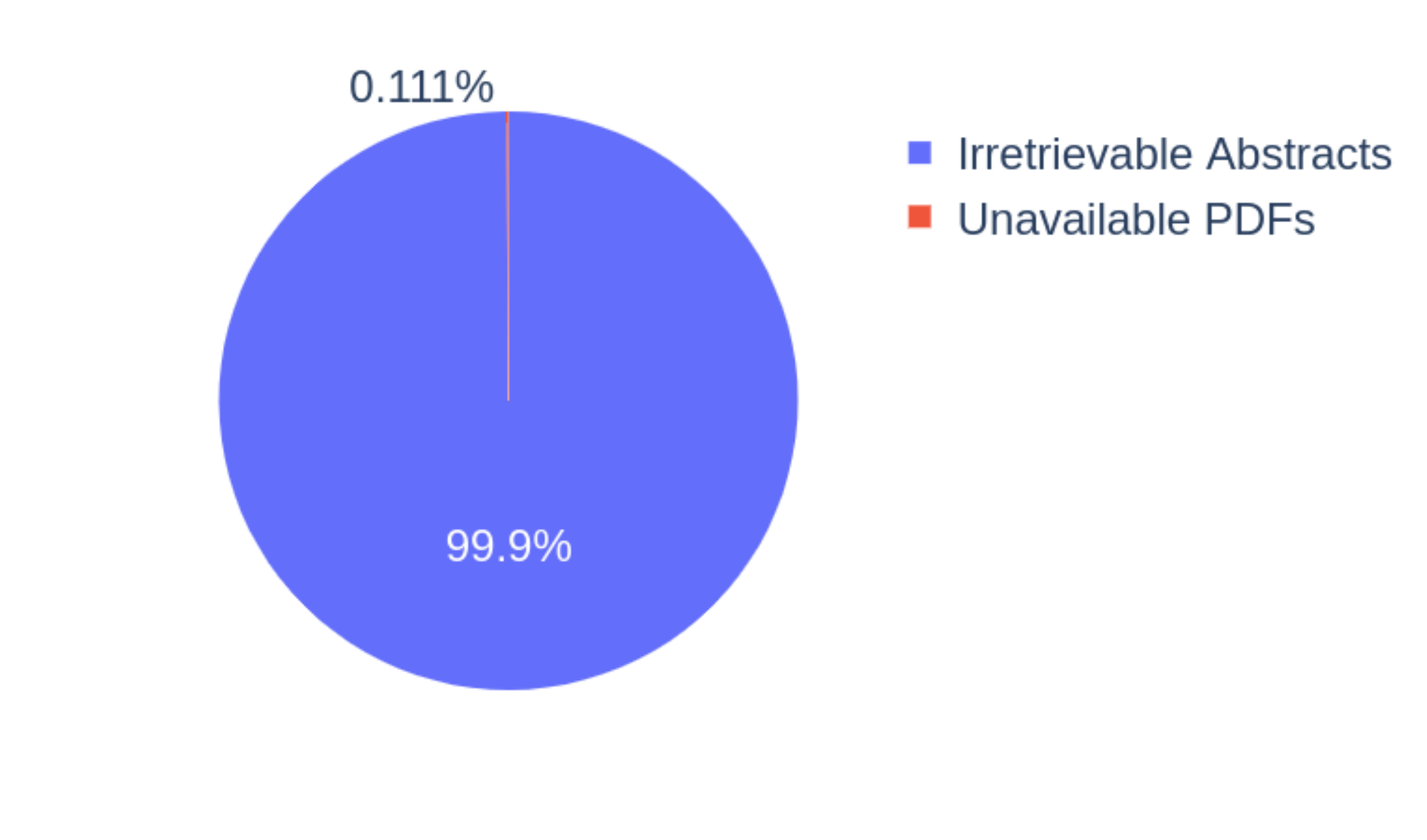}
  \end{subfigure}
  \begin{subfigure}[b]{0.4\textwidth}
    \includegraphics[width=\textwidth]{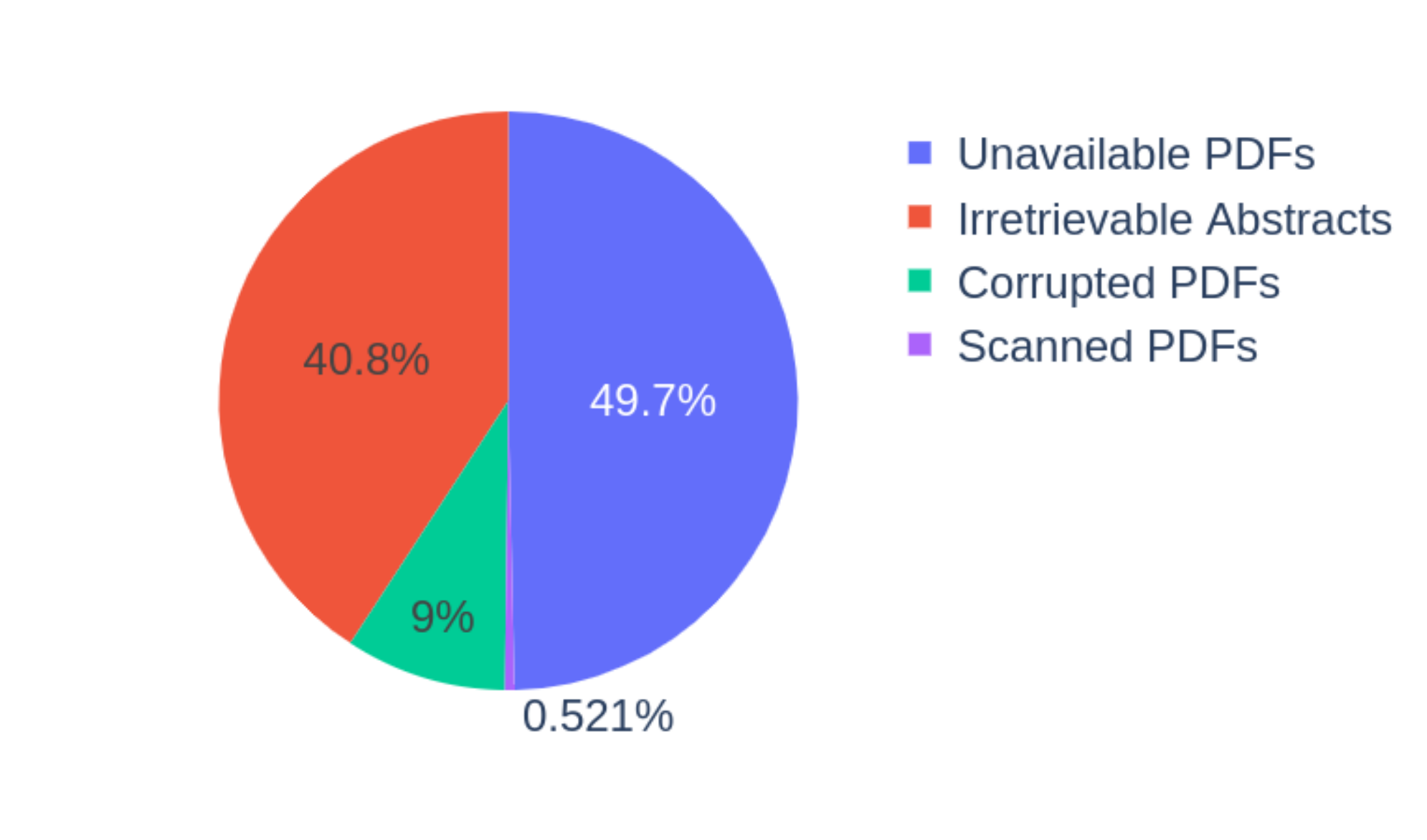}
  \end{subfigure}
\caption{Distribution of failure types in arXiv-Lay (top) and PubMed-Lay (bottom).}
\label{supp-fig:details-lost-docs}
\end{figure}

Figure~\ref{supp-fig:details-lost-docs} provides details on the amount of original documents lost in the process of augmenting arXiv and PubMed with layout/visual information. We observe four types of failures, and provide numbers for each type: 

\begin{itemize}
    \item The link to the document's PDF file is not provided (\textit{Unavailable PDF});
    \item The PDF file is corrupted (i.e., cannot be opened) (\textit{Corrupted PDF});
    \item The document is not digital-born, making it impossible to parse it with PDF parsing tools (\textit{ Scanned PDF});
    \item The document's abstract cannot be found in the PDF (\textit{Irretrievable Abstract}).
\end{itemize}

\subsection{KoreaScience -- Extraction Rule}
\label{supp-sec:koreascience-extraction}

\begin{figure}[ht]
\centering
    \includegraphics[width=0.49\textwidth]{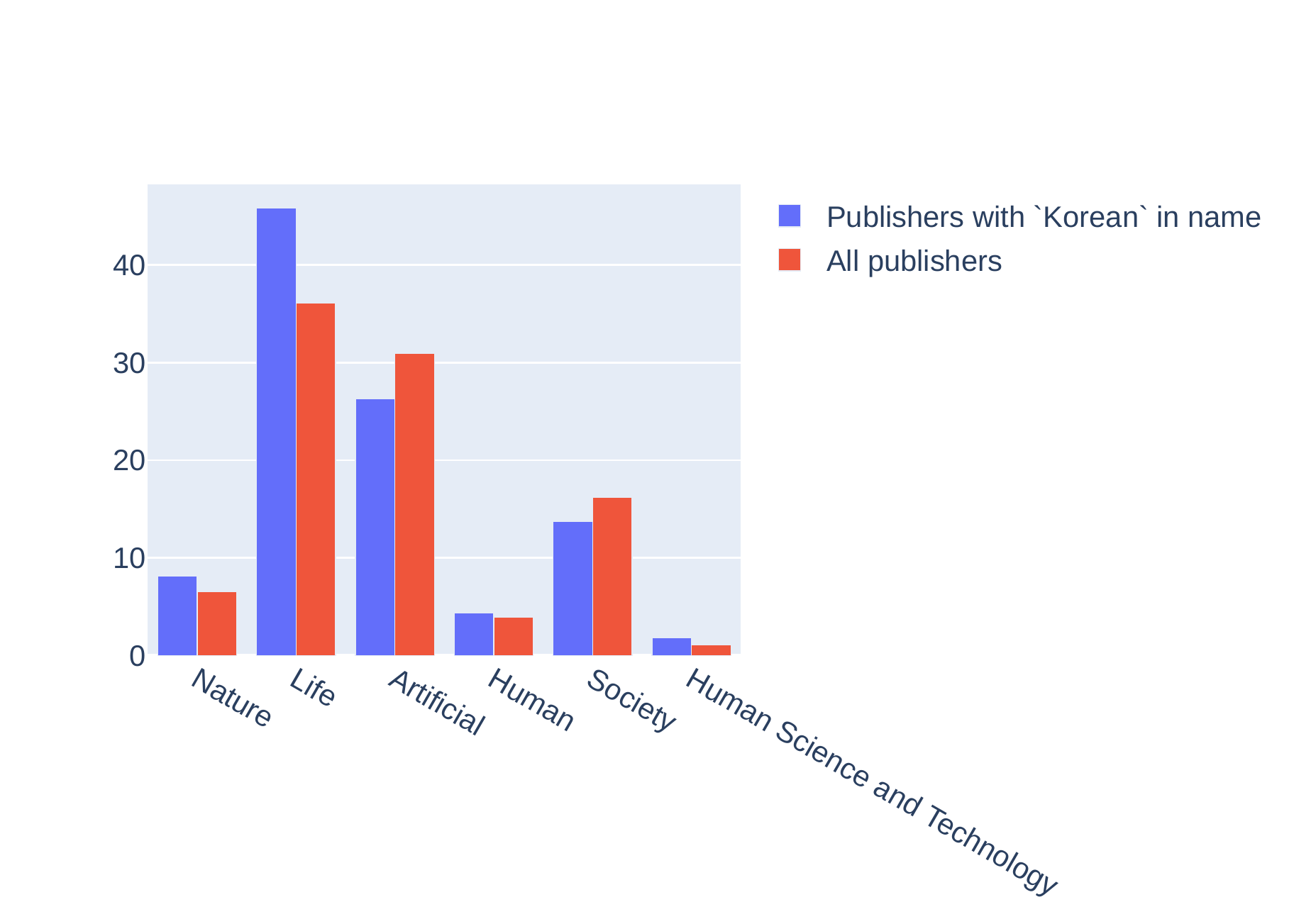}
  \caption{Distribution of topics covered by all publishers (red) vs distribution of topics covered by publishers whose name contains the word \textit{Korean} (blue).}
  \label{supp-fig:distr-koreascience-topics}
\end{figure}

Korean documents in KoreaScience are extracted by restricting search results to documents containing the word "Korean" in the publisher's name. We show that this rule does not bias the sample towards a specific research area. We compute the distribution of topics covered by all publishers, and compare it to the distribution of topics covered by publishers whose name contains the word \textit{Korean}. Figure~\ref{supp-fig:distr-koreascience-topics} shows that the distribution obtained using our rule remains roughly the same as the original.

\subsection{Samples}

\begin{figure*}[ht]
  \begin{subfigure}[b]{0.45\textwidth}
    \centering
    \includegraphics[width=0.6\textwidth]{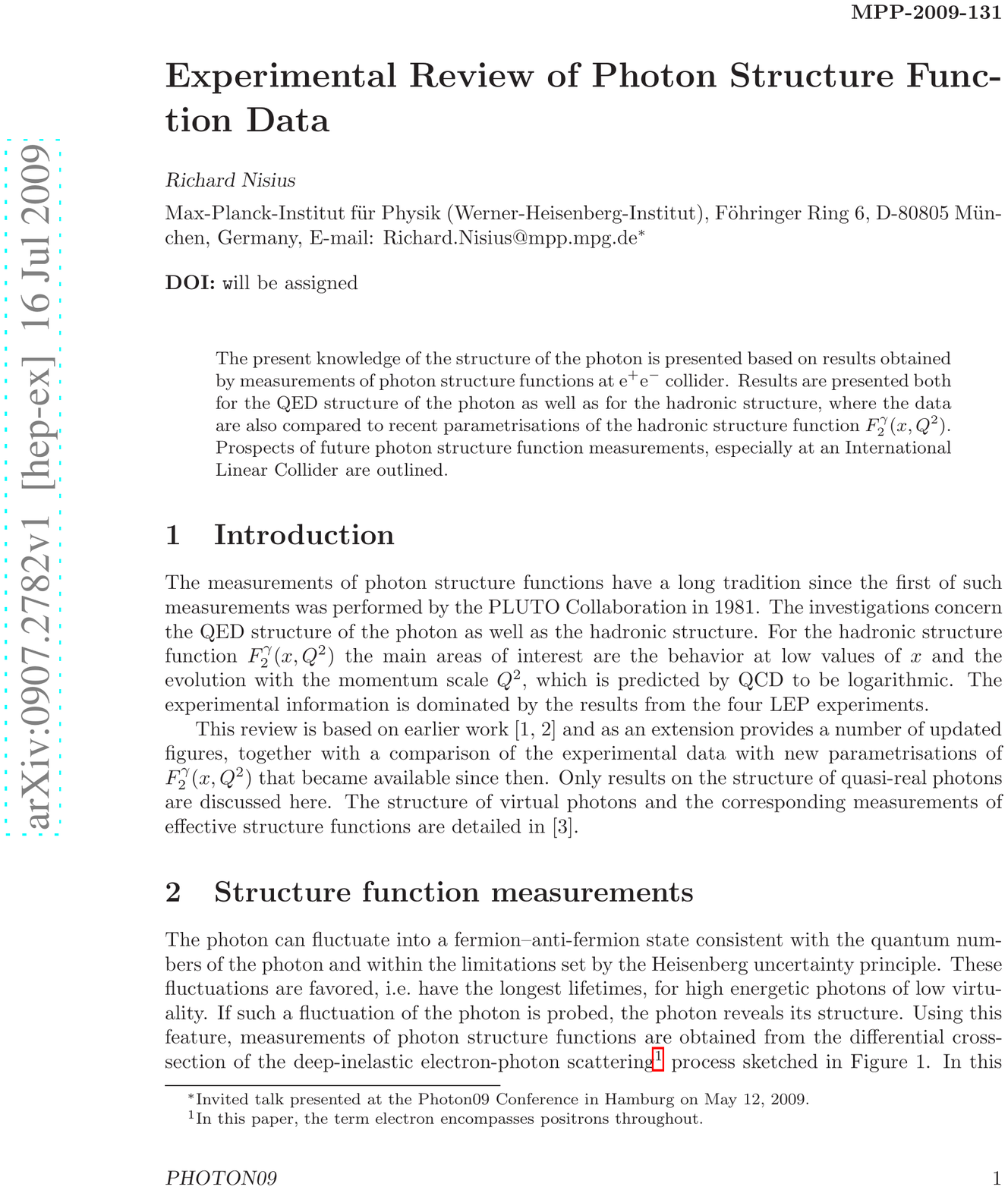}
    \caption{arXiv-Lay}
  \end{subfigure}
  \begin{subfigure}[b]{0.45\textwidth}
    \centering
    \includegraphics[width=0.6\textwidth]{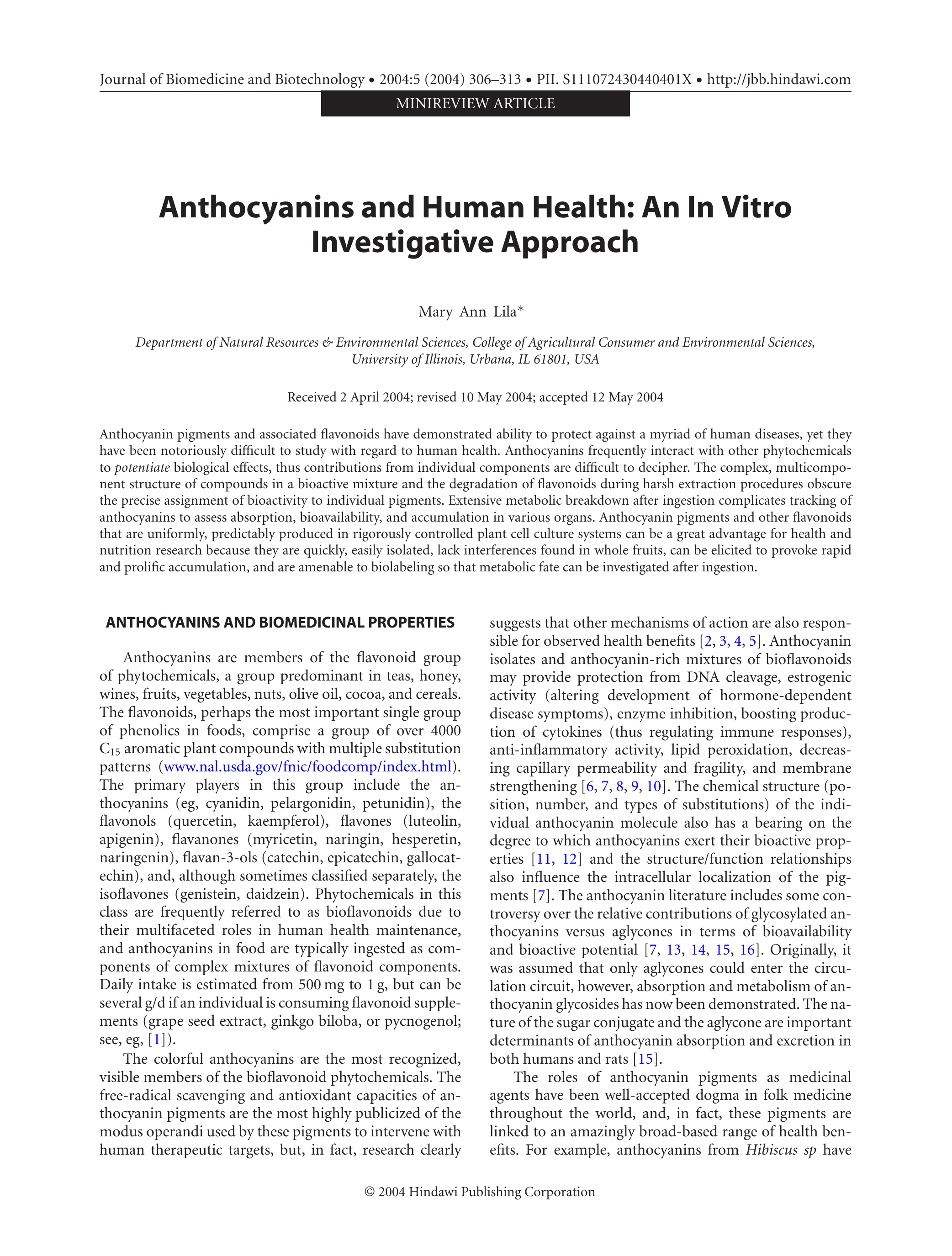}
    \caption{PubMed-Lay}
  \end{subfigure}
  \begin{subfigure}[b]{0.45\textwidth}
    \centering
    \includegraphics[width=0.6\textwidth]{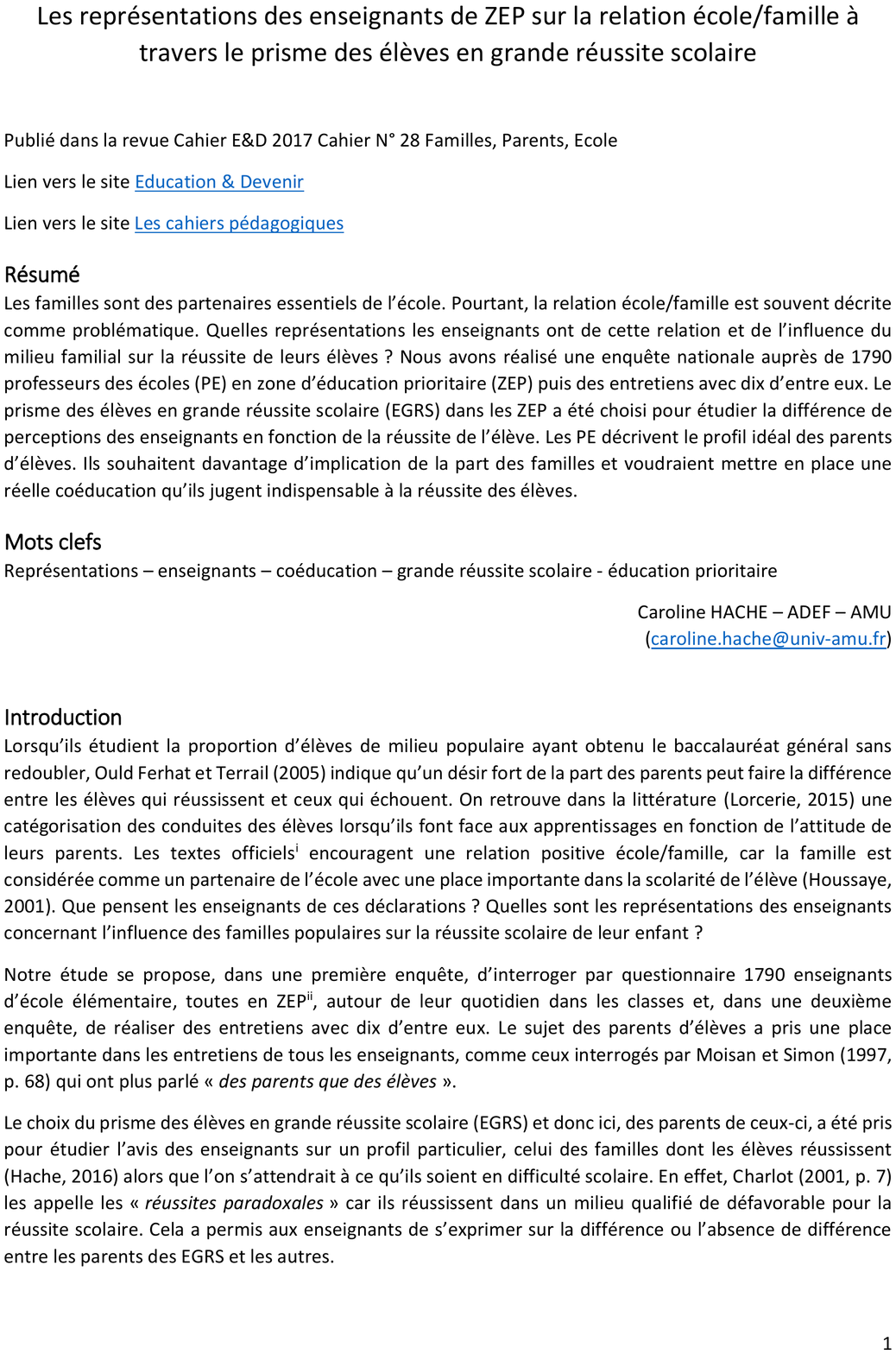}
    \caption{HAL}
  \end{subfigure}
  \begin{subfigure}[b]{0.45\textwidth}
    \centering
    \includegraphics[width=0.6\textwidth]{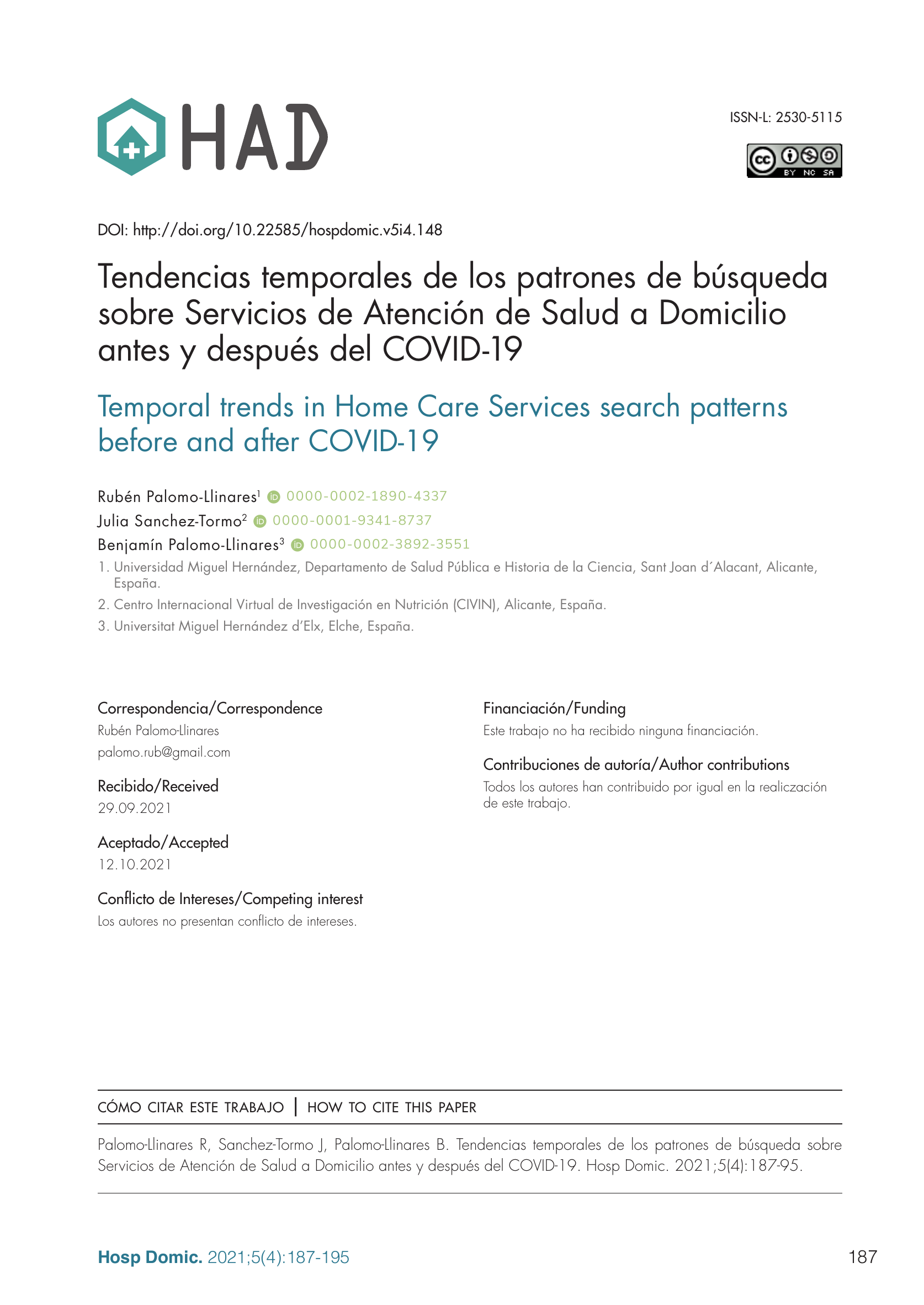}
    \caption{SciELO-ES}
  \end{subfigure}
  \begin{subfigure}[b]{0.45\textwidth}
    \centering
    \includegraphics[width=0.6\textwidth]{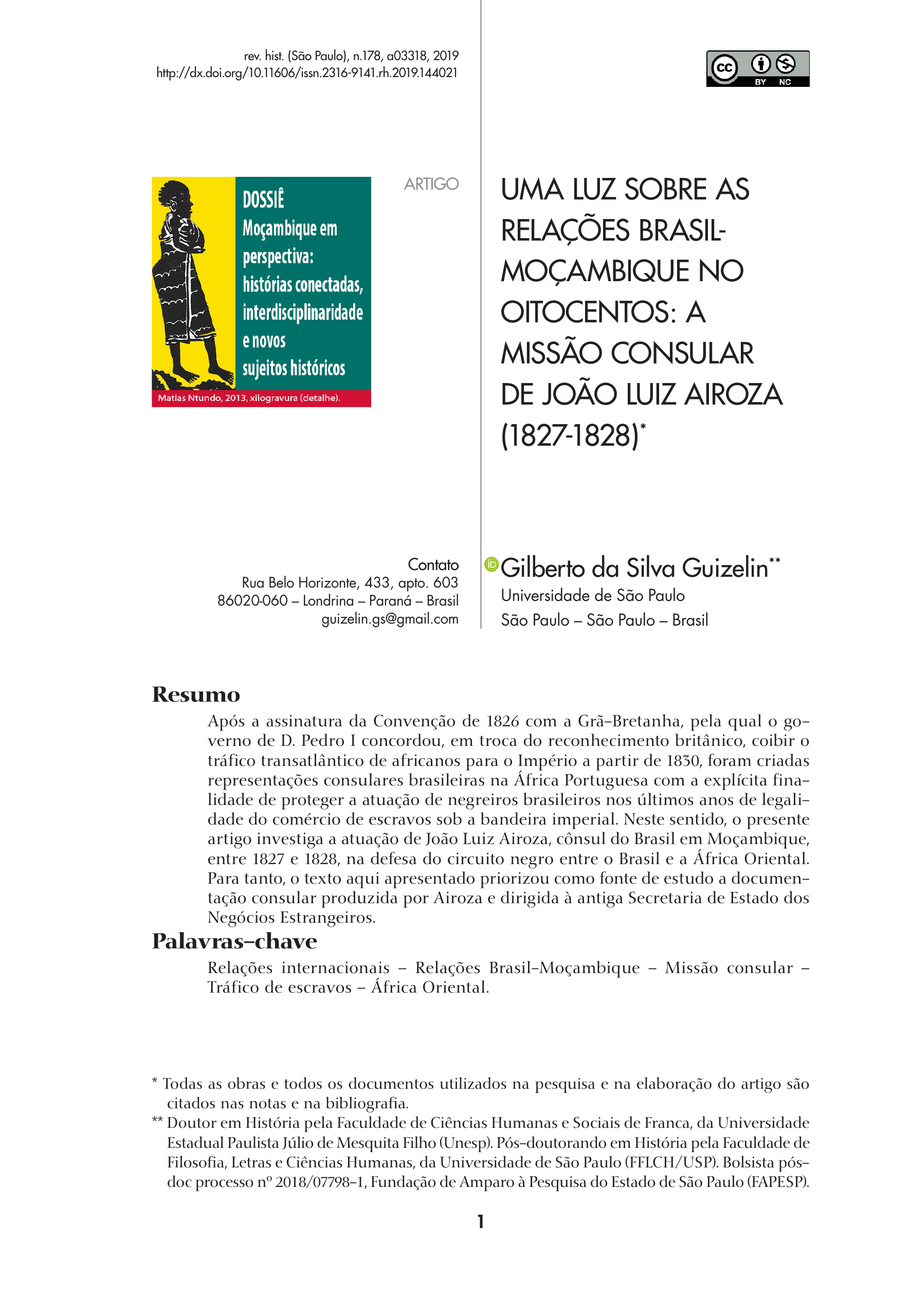}
    \caption{SciELO-PT}
  \end{subfigure}
  \hspace{1.5cm}
  \begin{subfigure}[b]{0.45\textwidth}
    \centering
    \includegraphics[width=0.6\textwidth]{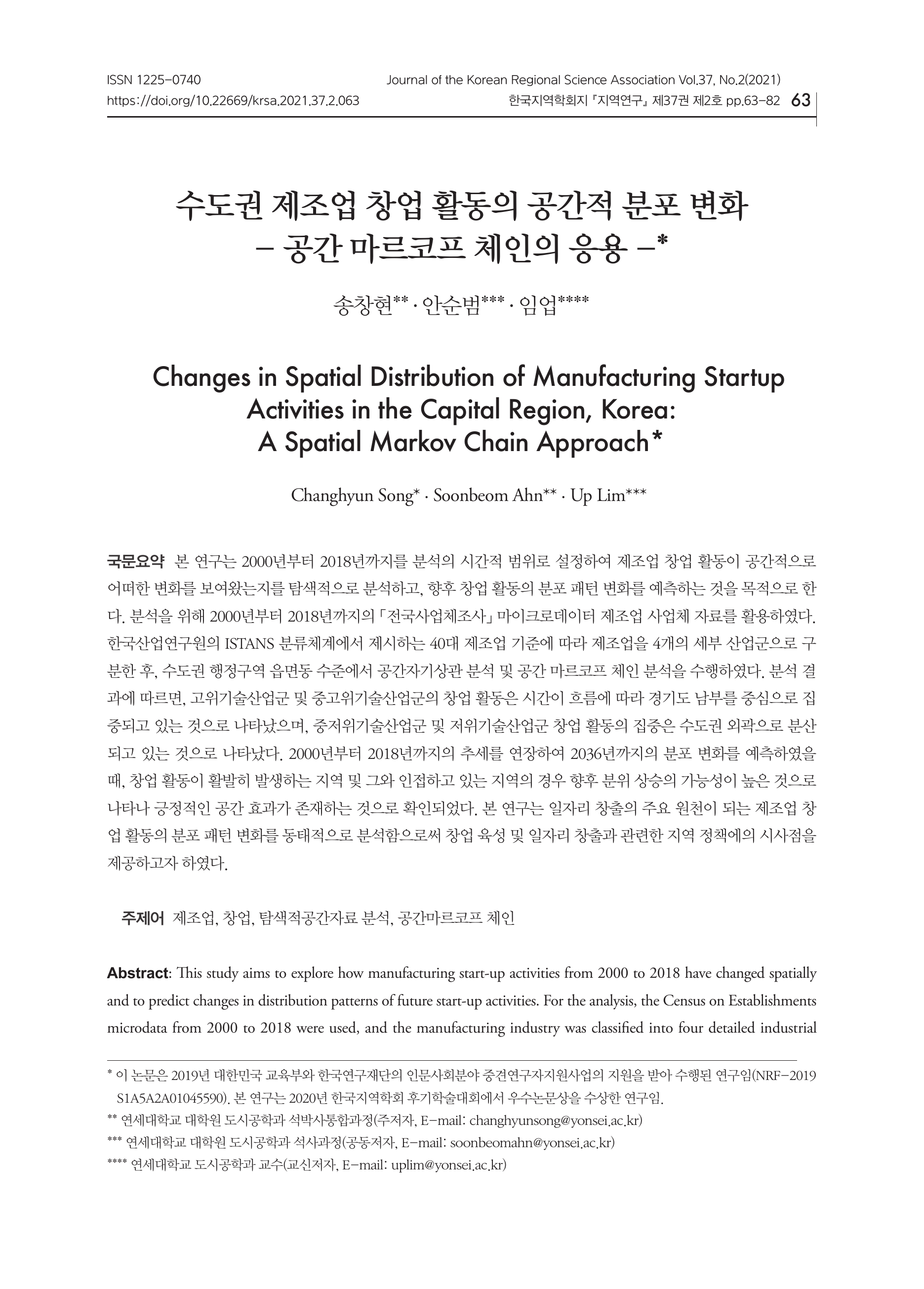}
    \caption{KoreaScience}
  \end{subfigure}
\caption{Samples from each dataset.}
\label{fig:samples-all}
\end{figure*}

We provide samples of documents from each dataset in Figure~\ref{fig:samples-all}.

\subsection{Datasets Statistics}
\label{supp-sec:datasets-statistics}

\begin{figure}[H]
    \includegraphics[width=0.5\textwidth]{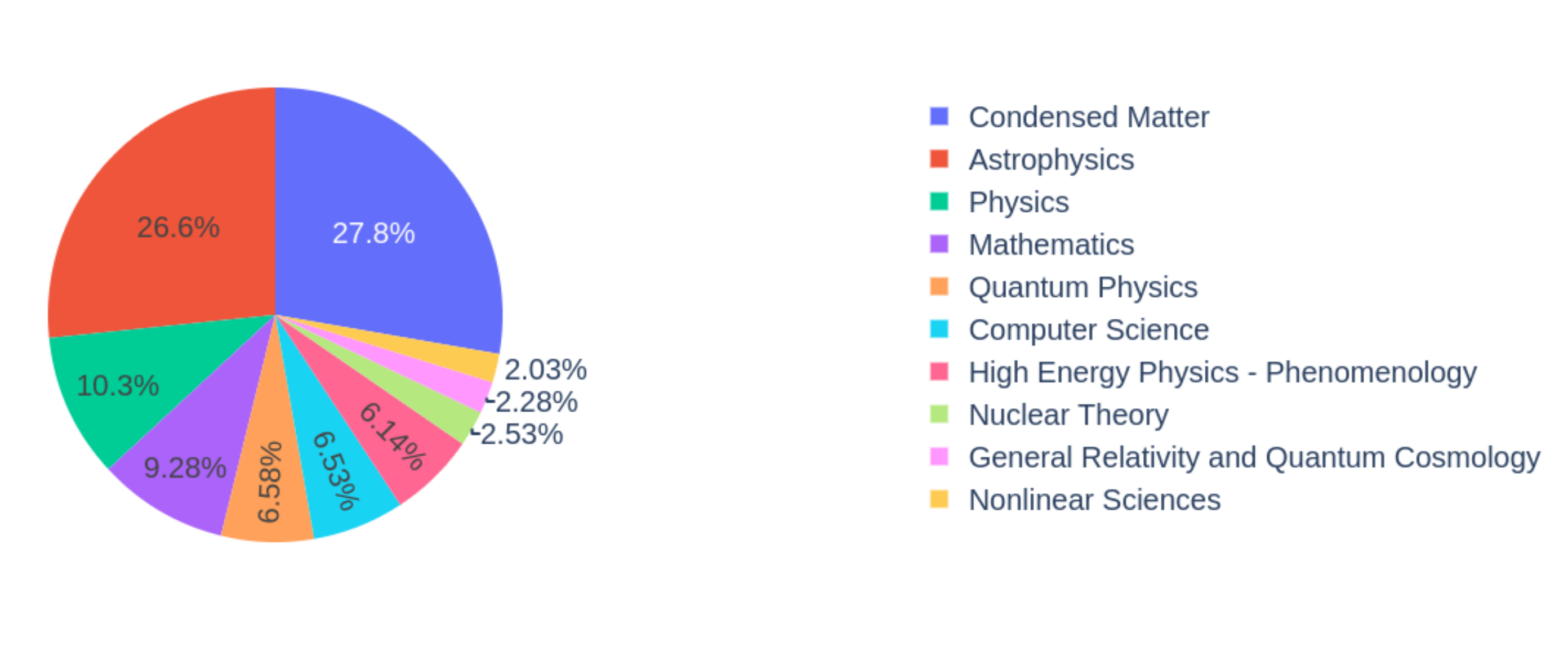}
  \caption{Distribution of research areas in arXiv-Lay.}
  \label{supp-fig:arxiv-research-areas}
\end{figure}

\begin{figure}[H]
    \includegraphics[width=0.5\textwidth]{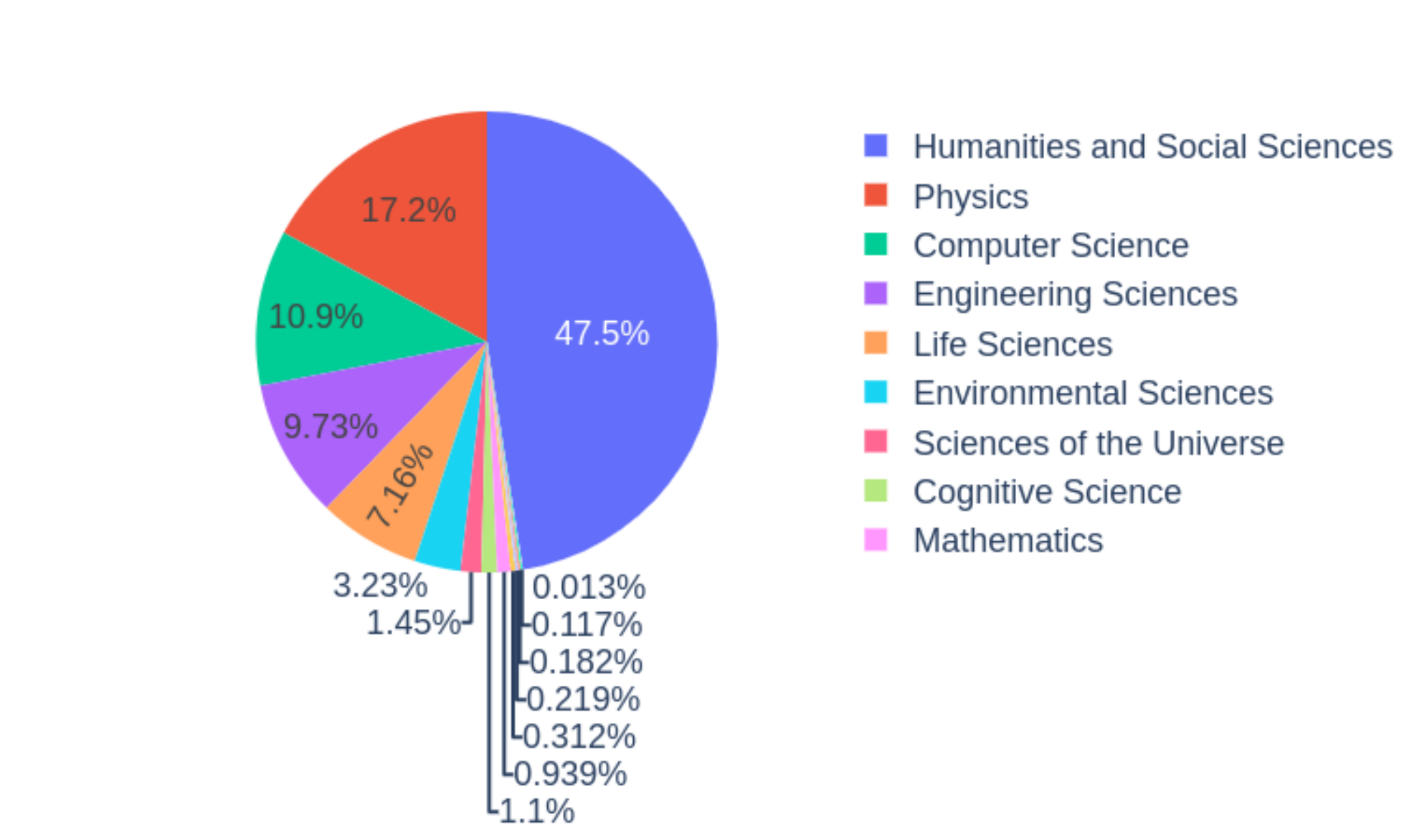}
  \caption{Distribution of research areas in HAL.}
  \label{supp-fig:hal-research-areas}
\end{figure}

The distribution of research areas in arXiv-Lay and HAL are provided in Figures~\ref{supp-fig:arxiv-research-areas} and ~\ref{supp-fig:hal-research-areas}, respectively. Such distributions are not available for the other datasets, as we did not have access to topic information during extraction.

\section{Experiments}

\subsection{Implementation Details}
\label{supp-sec:implementation-details}

Models were implemented in Python using PyTorch \citep{paszke2017automatic} and Hugging Face \citep{wolf2019huggingface} librairies. In all experiments, we use Adafactor \citep{shazeer2018adafactor}, a stochastic optimization method based on Adam \citep{kingma2014adam} that reduces memory usage while retaining the empirical benefits of adaptivity. We set a learning rate warmup over the first 10\% steps -- except on arXiv-Lay where it is set to 10k consistently with \citet{zaheer2020big}, and use a square root decay of the learning rate. All our experiments have been run on four Nvidia V100 with 32GB each. 

\section{Results}

\subsection{Detailed Results}
\label{supp-sec:detailed-results}

\begin{table*}[ht]
\centering
\small
\begin{tabular}{c*{12}{l}}
\toprule
\multicolumn{1}{c}{\multirow{2}{*}{\textbf{Model}}} & & \multicolumn{3}{c}{\textbf{arXiv / arXiv-Lay}}  &                                               & \multicolumn{3}{c}{\textbf{PubMed / PubMed-Lay}} \\ \cline{3-5} \cline{7-9} 
\multicolumn{1}{l}{}               &  & \multicolumn{1}{c}{R-1} & \multicolumn{1}{c}{R-2} & \multicolumn{1}{c}{R-L} & & R-1     & R-2     & R-L \\ 
\midrule
\rowcolor{Gray} Pegasus \citep{zhang2020pegasus}             &  & 44.21 & 16.95 & 38.83 & & 45.97 & 20.15 & 41.34 \\  
\rowcolor{Gray} BigBird-Pegasus \citep{zaheer2020big}        &  & 46.63 & 19.02 & 41.77 & & 46.32 & 20.65 & 42.33 \\  
\hline
T5 \citep{raffel2019exploring}    & & 42.79 & 15.98 & 37.90 & & 42.88 & 17.58	& 39.23 \\
LED \citep{beltagy2020longformer} & & 45.41 & 18.14 & 40.74 &	& 45.28	& 19.86	& 41.54 \\
LED+Layout                                          & & 45.51 & 18.55 & 40.96 &	& 45.41	& 19.74	& 41.83 \\
MBART                                               & & 37.64 &	13.29 & 33.49 & & 41.19	& 16.04	& 37.47 \\
Pegasus                 & & 43.81	            & 17.27	            & 39.07          & & 43.52          &  17.96          & 39.75  \\
Pegasus+Layout          & & 44.10	            & 17.01	            & 39.25          & & 43.59          &  18.24          & 39.85  \\
BigBird-Pegasus         & & 44.43	            & 17.74	            & 39.59          & & 44.80          &  19.32          & 41.09  \\
BigBird-Pegasus+Layout  & & \textbf{46.02}	& \textbf{18.95}	& \textbf{41.15} & & \textbf{45.69} & \textbf{20.38} & \textbf{42.05} \\
\bottomrule
      
\end{tabular}
\caption{ROUGE scores on arXiv-Lay and PubMed-Lay. Reported results obtained by Pegasus and BigBird-Pegasus on the original arXiv and PubMed are reported with a gray background. The best results obtained on arXiv-Lay and PubMed-Lay are denoted in bold.}
\label{supp-tab:results-arxiv-pubmed}
\end{table*}

\begin{table}[H]
\centering
\small
\begin{tabular}{clllllll}
\toprule

\textbf{Model} & \textbf{R-1} & \textbf{R-2} & \textbf{R-L} \\ 
\midrule
MBART                 & 47.05 & 22.23 & 42.00 \\
MBART+Layout          & 46.65 & 21.96 & 41.67 \\
BigBird-MBART         & 49.85 & \textbf{25.71} & 45.04 \\
BigBird-MBART+Layout  & \textbf{49.99} & 25.20 & \textbf{45.20} \\
\bottomrule
                            
\end{tabular}
\caption{ROUGE scores on HAL. Best results are reported in bold.}
\label{tab:results-hal}
\end{table}

\begin{table*}[h]
\centering
\small
\begin{tabular}{cllllllll}
\toprule
\multirow{3}{*}{\textbf{Model}} & \multicolumn{3}{c}{\textbf{SciELO-ES}} & & \multicolumn{3}{c}{\textbf{SciELO-PT}} \\  \cline{2-4} \cline{6-8} 
 &  R-1 & R-2 & R-L & & R-1 & R-2 & R-L \\ 
\midrule
MBART                 & 41.04 & 15.65 & 36.55 & & 41.18 & 15.53 & 36.42 \\
MBART+Layout          & 42.27 & 15.73 & 37.47 & & 39.45 & 14.17 & 34.37 \\
BigBird-MBART         & 42.64 & 16.60 & 37.76 & & 44.85 & 18.70 & 39.63 \\
BigBird-MBART+Layout  & \textbf{45.64} & \textbf{19.33} & \textbf{40.71} & & \textbf{45.47} & \textbf{20.40} & \textbf{40.51} \\
\bottomrule
                            
\end{tabular}

\caption{ROUGE scores on the SciELO datasets. The best results are reported in bold.}
\label{supp-tab:results-scielo}
\end{table*}

\begin{table}[H]
\centering
\small
\begin{tabular}{clllllll}
\toprule

\textbf{Model} & \textbf{R-1} & \textbf{R-2} & \textbf{R-L} \\ 
\midrule
MBART                 & 17.33 & 7.70 & 16.94 \\
MBART+Layout          & 15.43 & 6.69 & 14.98 \\
BigBird-MBART         & 18.96 & 8.01 & 18.55 \\
BigBird-MBART+Layout  & \textbf{20.36} & \textbf{9.49} & \textbf{19.95} \\
\bottomrule
                            
\end{tabular}
\caption{ROUGE scores on KoreaScience. The best results are reported in bold.}
\label{supp-tab:results-koreascience}
\end{table}

\begin{table*}[ht]
\centering
\small
\resizebox{\linewidth}{!}{%
\begin{tabular}{lcccccc}
\toprule
                        & \textbf{LED} & \textbf{LED+Layout} & \textbf{Pegasus} & \textbf{Pegasus+Layout} & \textbf{BigBird-Pegasus} & \textbf{BigBird-Pegasus+Layout} \\  
\midrule
\textbf{T5}                     & 2.84 / 2.31 & 3.06 / 2.60 & 1.17 / 0.52 & 1.35 / 0.62 & 1.69 / 1.86 & 3.25 / 2.82\\ 
\textbf{LED}                    & -- & 0.22 / 0.29 & 1.67 / 1.79 & 1.49 / 1.69 & 1.15 / 0.45 & 0.41 / 0.51\\
\textbf{LED+Layout}             & -- & -- & 1.89 / 2.08 & 1.71 / 1.98 & 1.38 / 0.74 & 0.19 / 0.22 \\
\textbf{Pegasus}                & -- & -- & -- & 0.34 / 0.10 & 0.52 / 1.34 & 2.08 / 2.30 \\ 
\textbf{Pegasus+Layout}         & -- & -- & -- & -- & 0.34 / 1.24 & 1.90 / 2.20 \\
\textbf{BigBird-Pegasus}        & -- & -- & -- & -- & -- & 1.56 / 0.96 \\
\bottomrule
                            
\end{tabular}
}
\caption{Absolute ROUGE-L score differences between each pair of models, on arXiv-Lay/PubMed-Lay.}
\label{supp-tab:rouge-l-improvements}
\end{table*}

\subsection{Human Evaluation}
\label{supp-sec:human-eval}

Using the Streamlit\footnote{\url{https://streamlit.io/}} framework, we design and develop an interface to aid human evaluation of summarization models.\footnote{The code is publicly available at \url{https://anonymous.4open.science/r/loralay-eval-interface-C20D}.}

\begin{figure}[H]
    \hspace{1cm}
    \includegraphics[width=0.6\textwidth]{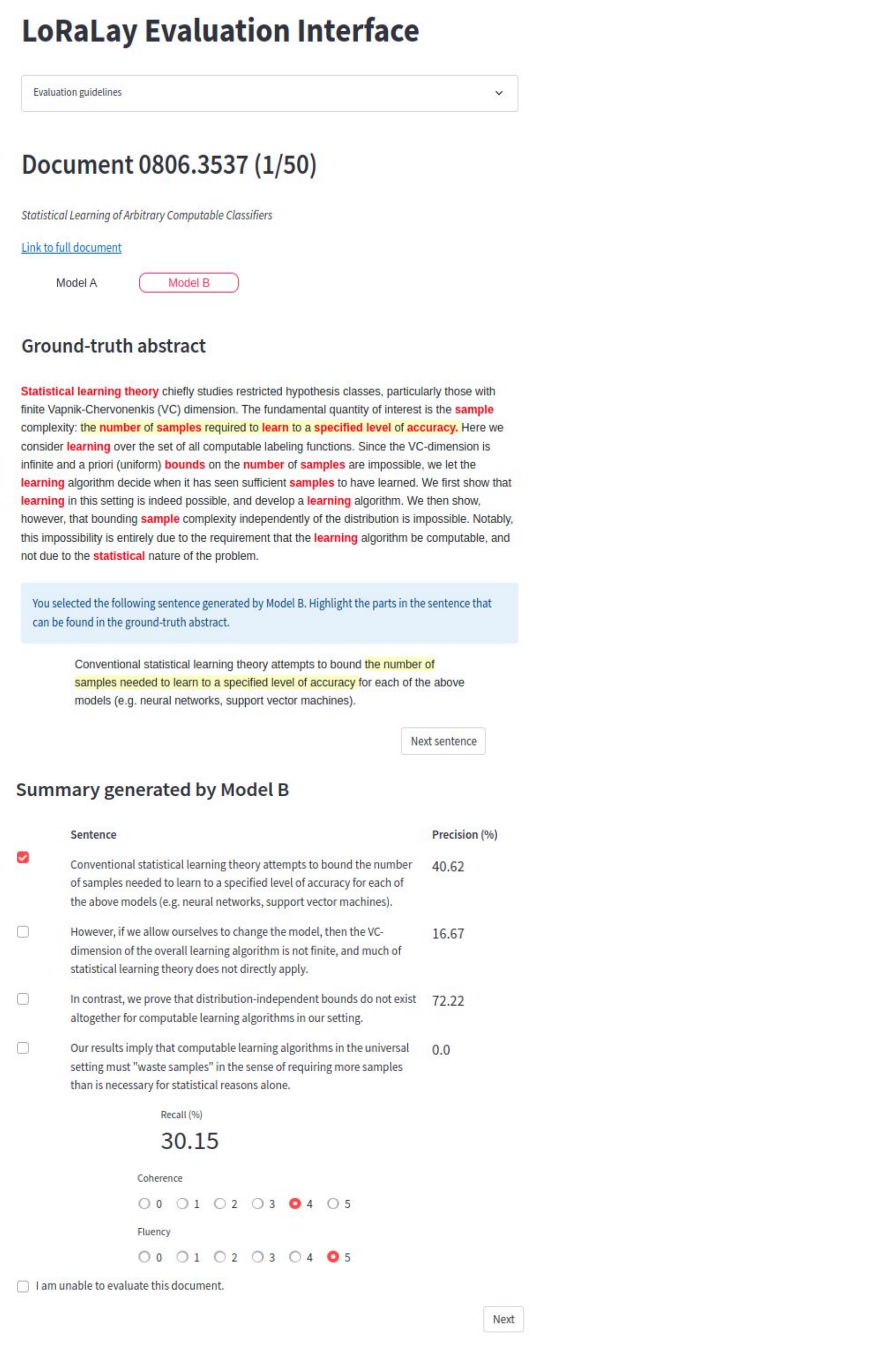}
  \caption{LoRaLay evaluation interface.}
  \label{supp-fig:loralay-eval-interface}
\end{figure}

\subsection{Analysis of the Impact of Layout}
\label{supp-sec:analysis-layout}

Table~\ref{supp-tab:quartiles} lists the quartiles computed from the distributions of article lengths, summary lengths, and variation in the height of bounding boxes, for arXiv-Lay and PubMed-Lay.

\begin{table*}[h]
\centering
\small
\begin{tabular}{crrrrrrrrr}
\toprule
\multirow{2}{*}{\textbf{Distribution}} & & \multicolumn{2}{c}{\textbf{Q1}} & & \multicolumn{2}{c}{\textbf{Q2}} & & \multicolumn{2}{c}{\textbf{Q3}} \\ \cline{3-4} \cline{6-7} \cline {9-10} 
               & & arXiv-Lay & PubMed-Lay & & arXiv-Lay & PubMed-Lay & & arXiv-Lay & PubMed-Lay \\ 
\midrule
Article Length                  & &    6,226 &    3,513 & &   9,142 &   5,557 & &   13,190 & 8,036 \\
Summary Length                  & &      119 &      130 & &     159 &     182 & &      202 &   247 \\  
$\sigma$ of bounding box height & &     3.37 &     1.34 & &    3.98 &    1.73 & &     4.70 &  2.28 \\
\bottomrule
      
\end{tabular}
\caption{Quartiles calculated from the distributions of article lengths, summary lengths, and variation in the height of bounding boxes, for arXiv-Lay and PubMed-Lay.}
\label{supp-tab:quartiles}
\end{table*}

\end{document}